%% file: main.tex
\pdfoutput=1
 \documentclass[twoside]{article}

\usepackage[accepted]{aistats2025}
\usepackage[round]{natbib}

\usepackage{macros}
\usepackage{amsmath}
\usepackage{mathtools}

\usepackage{hyperref}       
\hypersetup{
  colorlinks   = true, 
  urlcolor     = blue, 
  linkcolor    = blue, 
  citecolor   = blue 
}
\usepackage{url}            
\usepackage{booktabs}       
\usepackage{amsfonts}       
\usepackage{nicefrac}       
\usepackage{microtype}      
\usepackage{xcolor}         

\usepackage{array} 
\usepackage{graphicx}
\usepackage{subfigure}
\usepackage{caption}
\usepackage{tikz}
\usetikzlibrary {arrows.meta}
\usetikzlibrary {patterns,patterns.meta}
\usetikzlibrary{topaths,calc}
\usetikzlibrary {graphs}
\usetikzlibrary {graphs.standard}
\usepackage{enumerate}

\usepackage{algorithm2e}
\usepackage{algpseudocode}

\usepackage[title,page]{appendix}

\usepackage{thmtools,thm-restate}

\begin{document}

\newcommand{\suppmu}{s_{\mu}}
\newcommand{\omegamu}{\boldsymbol{\omega}_{\mu}}

\newcommand{\confgraph}[1][\suppmu]{G_{#1}^\epsilon}
\newcommand{\confhyp}[1][\suppmu]{H_{#1}^\epsilon}
\newcommand{\cliquehyp}[1][\suppmu]{C_{#1}^\epsilon}

\newcommand{\vecones}[1][n]{\bold 1_{#1}}
\newcommand{\unifvec}[1][n]{\frac{1}{#1}\vecones[#1]}

\renewcommand\vec{\boldsymbol}
\newcommand\separable{$\epsilon$-separable\xspace}
\newcommand\separability{$\epsilon$-separability\xspace}

%

%

\runningauthor{L. Gnecco Heredia, M. Sammut, M. Pydi, R. Pinot, B. Negrevergne, Y. Chevaleyre}

\twocolumn[

\aistatstitle{Unveiling the Role of Randomization in Multiclass Adversarial Classification: Insights from Graph Theory}

\aistatsauthor{Lucas Gnecco Heredia${}^{1}$, Matteo Sammut${}^{1}$, Muni Sreenivas Pydi${}^{1}$}
\aistatsauthor{Rafael Pinot${}^{2}$, Benjamin Negrevergne${}^{1}$, Yann Chevaleyre${}^{1}$}
\aistatsaddress{${}^1$ LAMSADE, Université Paris Dauphine - PSL, CNRS, Paris, France. \\
${}^2$ Laboratoire de Probabilités, Statistique et Modélisation, Sorbonne Université, Paris, France.} ]

\begin{abstract}

Randomization as a mean to improve the adversarial robustness of machine learning models has recently attracted significant attention. Unfortunately, much of the theoretical analysis so far has focused on binary classification, providing only limited insights into the more complex multiclass setting. In this paper, we take a step toward closing this gap by drawing inspiration from the field of graph theory. Our analysis focuses on discrete data distributions, allowing us to cast the adversarial risk minimization problems within the well-established framework of set packing problems. By doing so, we are able to identify three structural conditions on the support of the data distribution that are necessary for randomization to improve robustness. Furthermore, we are able to construct several data distributions where (contrarily to binary classification) switching from a deterministic to a randomized solution significantly reduces the optimal adversarial risk. These findings highlight the crucial role randomization can play in enhancing robustness to adversarial attacks in multiclass classification.

\end{abstract}


\vspace{-0.4cm}
\section{INTRODUCTION}
\label{sec:intro}

Modern machine learning models such as neural networks are highly vulnerable to small (imperceptible) adversarial perturbations~\cite{biggio2013evasion,szegedy2014intriguing}. Over the past decade, significant efforts have been made to develop strong attacks~\cite{goodfellow2014explaining,kurakin2016adversarial,carlini2017adversarial,croce2020reliable}, practical defense mechanisms~\cite{madry2017towards, moosavi2019robustness,cohen2019randomizedsmoothing,croce2020reliable,salman2019provably}, and advance the theoretical understanding of this phenomenon~\cite{awasthi2021existence_extended,pydi2023many,trillos2023existence,pydi2023many,meunier2022consistency,frank2024existence}. Among existing defense mechanisms, a prominent class known as \emph{randomized defenses} aims to enhance model robustness through the use of randomization. Initially explored in empirical studies~\cite{xie2017mitigating,dhillon2018stochastic,panousis2021local}, this approach has since gained significant attention in theoretical research~\cite{pinot2019theoretical, pinot2020randomization, pinot2022robustness, meunier2021mixed, gnecco2024role, yang2022rethinking, huang2022adversarial}. 

While the study of randomized defenses remains an open area of research, the extensive literature on adversarial example theory and on the specific role of randomization has provided valuable insights 
into how well randomized defenses might work, particularly in binary classification.
Notably, recent work~\cite{gnecco2024role} has shown that if the hypothesis class is sufficiently rich (e.g., when considering the set of all Borel measurable functions) then randomized strategies cannot enhance the robustness of the optimal classifier in binary classification. A similar result holds for Lebesgue measurable functions, when combining results from~\cite{meunier2021mixed} and~\cite{pydi2023many}.

The case of multiclass classification however remains largely understudied and missunderstood. One might assume that existing results in the binary classification framework would naturally extend to the multiclass setting, as is often the case in standard classification. However, perhaps unsurprisingly, this does not hold in the adversarial classification context. Recent work~\cite{trillos2023multimarginal,dai2024characterizing} has indeed presented a counterexample demonstrating that, in the multiclass setting, there exist simple data distributions with finite support where randomization improves the robustness of the optimal classifier, even when considering very rich hypothesis classes. The proof of this counterexample heavily relies on a symmetry argument, raising several follow-up questions about whether this example can be generalized. Specifically, we are interested in addressing the following question:

\textbf{Can we characterize the data distributions for which randomization enhances the adversarial robustness 
in multiclass classification?}

In this paper, we take a step toward providing a principled answer to this question. 
Specifically, we study finite discrete data distributions and analyze the value of the randomization gap (\textit{i.e.}, the difference between the optimal adversarial risk with and without randomization) for Borel measurable functions.\footnote{Since we focus on discrete distributions, any well-defined classifier is Borel measurable.}
To analyze the data distributions, we first present (in Section~\ref{sec: computing the randomization gap using graph theory}) a graph-theoretical characterization of their intrinsic vulnerabilities using the concepts of conflict graphs and conflict hypergraphs, first introduced in~\cite{dai2024characterizing}. We then map the computation of their randomization gap to two well-known graph-theoretical problems: set packing and fractional set packing. This approach enables us to derive two important sets of results that characterize the randomization gap based on specific structural properties of the data distribution we consider. Our main contributions are as follows. 

\textbf{Contribution 1: Identification of necessary structures for a positive randomization gap.} We demonstrate that the existence of a positive randomization gap (which indicates that randomization enhances robustness) in the multiclass adversarial classification problem is intrinsically linked to the presence of specific structural characteristics in the conflict hypergraph associated with the data distribution we study. Specifically, we show that a data distribution cannot exhibit a positive randomization gap unless its corresponding conflict hypergraph possesses one of the following features: it contains a hole, an anti-hole, or it misses to include all the cliques of its skeleton (as detailed in Section~\ref{sec: sufficient conditions}). This finding provides a necessary condition for a data distribution to exhibit a positive randomization gap, and thereby helps us gain deeper insights into the conditions under which randomization may be beneficial for robust multiclass classification.

\textbf{Contribution 2: Improved counterexamples with arbitrarily large randomization gap.} While our first contribution provides valuable insights, it does not constitute a sufficient condition for a positive randomization gap. Hence, it does not fully characterize the set of data distributions for which randomization improves robustness. In fact, we expect (see Section~\ref{sec: sufficient conditions}) that it is impossible to craft a sufficient condition which could be checked on a discrete distribution in polynomial time.
Nevertheless, inspired by our structural conditions, we present (in Section~\ref{sec:rg is big}) several novel counterexamples that generalize and enhance the initial results from~\cite{trillos2023multimarginal,dai2024characterizing}. Specifically, we identify a variety of data distributions and existence results for which we can establish that the randomization gap approaches $1/2$. This finding implies that, under certain conditions, the effectiveness of randomization can be arbitrary good. 

Our work not only extends existing literature but also opens new avenues for exploring the relationship between data distribution structures and the effectiveness of randomized defenses in the multiclass setting.

\section{PRELIMINARIES}
\label{sec:preliminaries}

\textbf{Notations.} 
We denote by $\mathbb{R}_{+}$ the set of non-negative real numbers, \textit{i.e}., $[0, \infty)$. For a vector $\boldsymbol{u}$, we denote by $\boldsymbol{u}^{(j)}$ the $j$-th component of $\vec{u}$. For a positive integer $K$, we use the notation $\left[K \right]$ to refer to the set $\{1, \dots, K\}$. We further denoted by $\Delta^K$, the \textit{probability simplex} in $\mathbb{R}^K$, \textit{i.e.}, $\Delta^K := \{ \vec{u} \in \mathbb{R}_+^K \mid  \sum_{i=1}^K \vec{u}^{(i)} =1\}$. 
We also denote $\vecones[n]$ the vector in $\mathbb{R}^n$ with all components one.

Additionally, for any space $\mathcal{Z}$, let $\mathcal{P}(\mathcal{Z})$ represent the set of finite discrete distributions, \textit{i.e.}, probability measures with finite support on $\mathcal{Z}$. A distribution $\mu \in \mathcal{P}(\mathcal{Z})$ is characterized by its support $s_\mu=\left( z_1, \dots, z_n \right)$ of size $n$, and its \textit{probability vector} $\omegamu \in \Delta^n$, which assigns a mass $\omegamu^{(i)}$ to each point $z_i$ in the support.
Finally, for any discrete space $\mathcal{Z}$, we denote by $2^{\mathcal{Z}}$ the \textit{power set} of $\mathcal{Z}$, \textit{i.e.}, the set of all subsets of $\mathcal{Z}$.

\textbf{Adversarial classification: deterministic setting.} Let $\mathcal{X} \subseteq \mathbb{R}^d$, $\mathcal{Y} = [K]$ and $\mu \in \mathcal{P}(\mathcal{X} \times \mathcal{Y})$.  Solving a classification task in the deterministic setting involves finding a function  $f$ in the set of  functions $\mathcal{F}_{\mathrm{det}} \coloneqq \{ f : \mathcal{X} \rightarrow \mathcal{Y} \}$ that minimizes the expected error on $\mu$ (a.k.a. the {\em risk}). More formally, it involves solving the following optimization problem:
\begin{equation}
\label{eq:StandardClassification}
    \inf_{f \in \mathcal{F}_\mathrm{det}}    
    \mathbb{E}_{(x, y) \sim \mu} \left[ \mathds{1}\{f(x) \neq y \} \right].
\end{equation}
However, the classifiers that are approximate solutions to~\eqref{eq:StandardClassification} may be vulnerable to {\em adversarial examples}. Given a classifier $f  \in \mathcal{F}_\mathrm{det}$ and a data sample $(x,y) \in \mathcal{X} \times \mathcal{Y}$, an adversarial example is an input $x_\text{adv} \in \mathcal{X}$ that is perceptively  indistinguishable from $x$ but is misclassified by $f$, \textit{i.e.}, $f(x_\text{adv}) \neq y$. Note that although the notion of perceptibility is a complicated concept that depends on human biology, it is common to measure the magnitude of an adversarial perturbation using an  $\ell_p$ norm (with $p \in [1, \infty]$). Thus an adversarial example  $x_\text{adv}$ is considered  indistinguishable from $x$ when $\norm{x_\text{adv} - x}_p \leq \epsilon$ with $\epsilon$  chosen empirically. 
To account for the existence of adversarial examples, we now define the problem of {\em adversarial classification} as follows: 
\begin{equation}
\label{eq:min-risk-adv-det}
     \inf_{f \in \mathcal{F}_\mathrm{det}} \mathbb{E}_{(x, y) \sim \mu} \left[ \sup_{x_\text{adv} \in B_p(x, \epsilon)}  \mathds{1}\{f(x_\text{adv}) \neq y \} \right],
\end{equation}
where $B_p(x, \epsilon) \coloneqq \{ x_\text{adv} \in \mathcal{X} : \norm{x_\text{adv} - x}_p \leq \epsilon \}$. The value of \eqref{eq:min-risk-adv-det} is the called the {\em optimal adversarial risk for deterministic classifiers}, and we denote this risk by $\cal R_{\cal F_\mathrm{det}}^*(\mu,\epsilon)$ in the rest of this paper. 

\textbf{Adversarial classification: randomized setting.} In the randomized setting, the goal is identical, except we now consider a wider set of functions $\mathcal F_\mathrm{rand}$ that includes {\em randomized classifiers}.  Formally $\mathcal F_\mathrm{rand}$ can be defined as follows: 
\begin{equation}
   \mathcal{F}_\mathrm{rand} \coloneqq \{ f : \mathcal{X} \rightarrow \mathcal{P}(\mathcal{Y})\}.
\end{equation}
There are several differences between the randomized and the deterministic setting. First, if $f$ is a randomized classifier from  $\mathcal{F}_\mathrm{rand}$ and $x \in \mathcal{X}$ is an arbitrary input, then obtaining a class label for $x$ using $f$ now requires sampling a class label $z \sim f(x)$. Second, the error made by a randomized classifier needs to be computed in expectation with respect to this sampling procedure. By analogy with the deterministic setting, we can thus define the {\em adversarial classification problem} for {\em randomized classifiers} as follows:
\begin{equation}
\label{eq:min-risk-adv-rand}
    \inf_{f \in \mathcal{F}_\mathrm{rand}} \mathbb{E}_{(x, y) \sim \mu} \left[ \sup_{x' \in B_p(x, \epsilon)} \hspace{-0.2cm} \mathbb{E}_{z \sim f(x')} \left[ \mathds{1}\{z \neq y \} \right] \right].
\end{equation}

The value of \eqref{eq:min-risk-adv-rand} is the {\em optimal adversarial risk for randomized classifiers}, which we denote $\cal R_{\cal F_\mathrm{rand}}^*(\mu,\epsilon)$. 

\textbf{Randomization gap.} Note that any deterministic classifier can be rewritten as a randomized one that only outputs Dirac measures. Hence, the set of deterministic classifiers $\mathcal{F}_\mathrm{det}$ can be conceptually considered as a subset of randomized classifiers $\mathcal{F}_\mathrm{rand}$. Accordingly, we always have that  $\mathcal{R}_{\cal F_{\mathrm{rand}}}^*(\mu,\epsilon) \le \mathcal{R}_{\cal F_{\mathrm{det}}}^*(\mu,\epsilon)$. In this paper, we are interested in determining the characteristics of $\mu$ that may lead this inequality to either become an equality or a strict inequality. In other words, we study the notion of \textit{randomization gap} of a distribution $\mu \in \mathcal{P}(\mathcal{X\times\mathcal{Y}})$, defined as:
\begin{equation} \label{eq: randomization gap definition}
\mathrm{rg}(\mu, \epsilon) \coloneqq \mathcal{R}_{\cal F_{\mathrm{det}}}^*(\mu,\epsilon) - \mathcal{R}_{\cal F_{\mathrm{rand}}}^*(\mu,\epsilon). 
\end{equation}
We offer a novel framework for understanding, and measuring the randomization gap, by focusing on finite discrete distributions. Specifically, we take inspiration from a reformulation of the adversarial risk minimization first introduced in~\cite{dai2024characterizing} to express the randomization gap as a graph theoretical problem, as we describe in the next section.

\section{COMPUTING THE RANDOMIZATION GAP USING GRAPH THEORETICAL TOOLS} \label{sec: computing the randomization gap using graph theory}

To compute the randomization gap of a distribution $\mu \in \mathcal{P}\left( \mathcal{X}\times\mathcal{Y} \right)$ using graph theoretical tools, we rely on the concept of {\em conflict hypergraph}, a representation that captures the intrinsic vulnerabilities of the optimal classifier on $\mu$, first introduced in~\cite{dai2024characterizing}. Using this representation, we map the adversarial risk minimization problems~\eqref{eq:min-risk-adv-det} and~\eqref{eq:min-risk-adv-rand} to well studied graph theoretical problems, known as set packing and fractional set packing. Before diving into the technical details, let us present some graph theoretical terminology.

\textbf{Graph theoretical terminology.} A simple undirected graph $G = (V, E)$ consists of a set of vertices $V$ and a set of edges $E \subseteq \{ e \in  2^V \text{ s.t. } \vert e\vert = 2\}$ denoting the connections between the vertices. 
We denote $|G|$ the number of vertices of $G$.
A \emph{hypergraph} $H = (V, \mathcal{E})$ is a generalization of a graph in which the hyperedges are subsets of $V$ and can thus contain more than two vertices. We call $k$-hyperedges all  hyperedges that connect exactly $k$ vertices. Note that a hypergraph that only contains $2$-hyperedges is a simple graph.

\subsection{Conflict hypergraph, conflict graph}

The notions of {\em conflict hypergraph} and {\em conflict graph} are both useful to characterize  intrinsic vulnerabilities of a finite discrete distribution $ \mu \in \mathcal{P}(\mathcal{X} \times \mathcal{Y}) $ to adversarial attacks. They rely on the concept of a {\em conflict}:  two points  $\left(x_1, y_1\right)$ and   $ (x_2, y_2)$ from the support of $\mu$ are said to be in conflict if they are both  within close range ($ \|x_1 - x_2\|_p \leq 2 \epsilon $) and belong to different classes ($ y_1 \neq y_2 $). Informally, configurations of two or more points in conflict are relevant to our problem because the adversarial risk has to be positive on at least one of these points, thus points involved in a conflict are vulnerable to adversarial attacks. To assess the vulnerability of 
a distribution $\mu$, we record all the conflicts as hyperedges in a hypergraph that has a vertex for each element of the support $\mu$. However, hypergraphs are sometime difficult to analyze, and thus it is also useful to consider the conflict graph of $\mu$, which only records pairwise conflicts between points. The conflict hypergraph, and the conflict graph, can be defined as:

\medskip

\makeatletter
\tikzset{
    dot diameter/.store in=\dot@diameter,
    dot diameter=3pt,
    dot spacing/.store in=\dot@spacing,
    dot spacing=10pt,
    dots/.style={
        line width=\dot@diameter,
        line cap=round,
        dash pattern=on 0pt off \dot@spacing
    }
}
\makeatother

\def\scalefig{0.35}
\def\radius{3}
\def\eps{\radius*0.77}
\def\epssmall{\radius*0.16}
\def\aconst{1.73205080757} 
\def\bconst{\radius*\aconst/2}
\def\cyclefactor{0.9}

\newcommand\figscale{0.25}
\begin{figure*}[ht]
    \centering
    \captionsetup{width=0.5\linewidth,labelformat=simple}
    \minipage{0.49\textwidth}
    \centering
      
    \begin{tikzpicture}[scale=\scalefig*0.85]
    
        \pgfmathsetmacro\angle{72*1}
        \pgfmathsetmacro\x{\radius*cos(\angle)*\cyclefactor}
        \pgfmathsetmacro\y{\radius*sin(\angle)*\cyclefactor}
    
        \fill[red] (\x, \y) circle (1mm) node[anchor=south west, xshift=-3pt, yshift=0pt] {$x_1$};
        \draw[black] (\x, \y) circle [radius=\eps];
    
        
        \pgfmathsetmacro\angle{72*2}
        \pgfmathsetmacro\x{\radius*cos(\angle)*\cyclefactor}
        \pgfmathsetmacro\y{\radius*sin(\angle)*\cyclefactor}
    
        \fill[blue] (\x, \y) circle (1mm) node[anchor=south east, xshift=3pt, yshift=0pt] {$x_2$};
        \draw[black] (\x, \y) circle [radius=\eps];
    
        
        \pgfmathsetmacro\angle{72*3}
        \pgfmathsetmacro\x{\radius*cos(\angle)*\cyclefactor}
        \pgfmathsetmacro\y{\radius*sin(\angle)*\cyclefactor}
    
        \fill[black!30!green] (\x, \y) circle (1mm) node[anchor=north east, xshift=3pt, yshift=1pt] {$x_3$};
        \draw[color=black] (\x, \y) circle [radius=\eps];
        
        
        \pgfmathsetmacro\angle{72*4}
        \pgfmathsetmacro\x{\radius*cos(\angle)*\cyclefactor}
        \pgfmathsetmacro\y{\radius*sin(\angle)*\cyclefactor}
    
        \fill[cyan] (\x, \y) circle (1mm) node[anchor=north, xshift=1pt, yshift=0pt] {$x_4$};
        \draw[black] (\x, \y) circle [radius=\eps];
    
        
        \pgfmathsetmacro\angle{72*5}
        \pgfmathsetmacro\x{\radius*cos(\angle)*\cyclefactor}
        \pgfmathsetmacro\y{\radius*sin(\angle)*\cyclefactor}
        \fill[purple] (\x, \y) circle (1mm) node[anchor=west] {$x_5$};
        \draw[black] (\x, \y) circle [radius=\eps];

    \end{tikzpicture}
    \hspace{2em}
    \begin{tikzpicture}[scale=\scalefig*0.8]

        \pgfmathsetmacro\angle{72*1}
        \pgfmathsetmacro\x{\radius*cos(\angle)}
        \pgfmathsetmacro\y{\radius*sin(\angle)}
        \coordinate (x1) at (\x, \y);
        \fill[red] (x1) circle (1.5mm) node[anchor=south west, xshift=3pt, yshift=4pt] {$x_1$};
        \draw[orange!70, fill=orange!70, fill opacity = 0.3] (\x, \y) circle [radius=\epssmall];
    
        \pgfmathsetmacro\angle{72*2}
        \pgfmathsetmacro\x{\radius*cos(\angle)}
        \pgfmathsetmacro\y{\radius*sin(\angle)}
        \coordinate (x2) at (\x, \y);
        \fill[blue] (x2) circle (1.5mm) node[anchor=south east, xshift=-4pt, yshift=4pt] {$x_2$};
        \draw[orange!70, fill=orange!70, fill opacity = 0.3] (\x, \y) circle [radius=\epssmall];
        
        \pgfmathsetmacro\angle{72*3}
        \pgfmathsetmacro\x{\radius*cos(\angle)}
        \pgfmathsetmacro\y{\radius*sin(\angle)}
        \coordinate (x3) at (\x, \y);
        \fill[black!30!green] (x3) circle (1.5mm) node[anchor=north east, xshift=-3pt, yshift=-4pt] {$x_3$};
        \draw[orange!70, fill=orange!70, fill opacity = 0.3] (\x, \y) circle [radius=\epssmall];
        
        \pgfmathsetmacro\angle{72*4}
        \pgfmathsetmacro\x{\radius*cos(\angle)}
        \pgfmathsetmacro\y{\radius*sin(\angle)}
        \coordinate (x4) at (\x, \y);
        \fill[cyan] (x4) circle (1.5mm) node[anchor=north, xshift=2pt, yshift=-9pt] {$x_4$};
        \draw[orange!70, fill=orange!70, fill opacity = 0.3] (\x, \y) circle [radius=\epssmall];

        \pgfmathsetmacro\angle{72*5}
        \pgfmathsetmacro\x{\radius*cos(\angle)}
        \pgfmathsetmacro\y{\radius*sin(\angle)}
        \coordinate (x5) at (\x, \y);
        \fill[purple] (x5) circle (1.5mm) node[anchor=west, xshift=7pt, yshift=3pt] {$x_5$};
        \draw[orange!70, fill=orange!70, fill opacity = 0.3] (\x, \y) circle [radius=\epssmall];
        
        
        \begin{scope}[fill opacity=0.1, color=orange, fill=orange!70]
        \filldraw[]         ($(x1)+(0,1.0)$) 
            to[out=170,in=15] ($(x1) + (-2.0,-0.5)$) 
            to[out=200,in=90] ($(x1) + (-4.2,-0.8)$) 
            to[out=270,in=215] ($(x1) + (-2,-1.6)$) 
            to[out=30,in=200] ($(x1) + (0,-1.1)$) 
            to[out=10,in=270] ($(x1) + (0.8, 0.)$) 
            to[out=90,in=0] ($(x1) + (0, 1.0)$) 
            ;
        
        \filldraw[]         ($(x2)+(0,1.3)$) 
            to[out=180,in=80] ($(x2) + (-0.4,-2.5)$) 
            to[out=250,in=180] ($(x2) + (-0.,-4.8)$)
            to[out=0,in=260] ($(x2) + (0.8,-1.7)$)
            to[out=75,in=0] ($(x2) + (0,1.3)$)
            ;
            
            \filldraw[]         ($(x3)+(-0.8,0.8)$) 
            to[out=215,in=135] ($(x3) + (-0.,-0.8)$) 
            to[out=315,in=180] ($(x3) + (2,-1.3)$) 
            to[out=10,in=200] ($(x3) + (4.4,-1.8)$)
            to[out=15,in=350] ($(x3) + (3.6,-0.4)$)
            to[out=180,in=0] ($(x3) + (1,0.1)$)
            to[out=180,in=45] ($(x3) + (-0.8,0.8)$)
            ;
            
            \filldraw[]         ($(x4)+(-0.7,-0.7)$) 
            to[out=300,in=260] ($(x4) + (1.8,1.5)$)
            to[out=80,in=290] ($(x4) + (2.9,3.4)$)
            to[out=135,in=45] ($(x4) + (1.5,3.2)$)
            to[out=220,in=30] ($(x4) + (0.,0.6)$)
            to[out=195,in=135] ($(x4) + (-0.7,-0.7)$)
            ;
            
            \filldraw[]         ($(x5)+(0.9,-0.9)$) 
            to[out=40,in=290] ($(x5) + (-1.,2.2)$)
            to[out=110,in=25] ($(x5) + (-2.5,3.5)$)
            to[out=200,in=135] ($(x5) + (-1, 0.3)$)
            to[out=320,in=215] ($(x5) + (0.9, -0.9)$)
            ;
        \end{scope}

        \path (x1) edge [color=black, dashed, dash pattern=on 2pt off 1pt] (x2);
        \path (x2) edge [color=black, dashed, dash pattern=on 2pt off 1pt] (x3);
        \path (x3) edge [color=black, dashed, dash pattern=on 2pt off 1pt] (x4);
        \path (x4) edge [color=black, dashed, dash pattern=on 2pt off 1pt] (x5);
        \path (x5) edge [color=black, dashed, dash pattern=on 2pt off 1pt] (x1);
    
    \end{tikzpicture}      
        \caption{}
        \label{fig: toy example c5}
    \endminipage \hfill\vline\hfill
    \minipage{0.48\textwidth}%
        \centering
        \def\epssmall{\radius/5}
    \begin{tikzpicture}[scale=\scalefig]
    
            \coordinate (x1) at (0,0);
            \fill[red] (x1) circle (1mm) node[anchor=north east] {$x_1$};
            \draw[black] (x1) circle [radius=\eps];
        
            \coordinate (x2) at (\radius,-\radius*0.5);
            \fill[blue] (x2) circle (1mm) node[anchor=north west] {$x_2$};
            \draw[black] (x2) circle [radius=\eps];
        
            \coordinate (x3) at (\radius/2,\bconst);
            \fill[black!30!green] (x3) circle (1mm) node[anchor=south] {$x_3$};
            \draw[color=black] (x3) circle [radius=\eps];

            \coordinate (x4) at (\radius*1.7,\bconst*1.1);
            \fill[purple] (x4) circle (1mm) node[anchor=south] {$x_4$};
            \draw[color=black] (x4) circle [radius=\eps];

    \end{tikzpicture}
    \hspace{2em}
    \begin{tikzpicture}[scale=\scalefig]

        \coordinate (x1) at (0,0);
        \fill[red] (x1) circle (1.5mm) node[anchor=north east, yshift=-9pt, xshift=-3pt] {$x_1$};
        \draw[orange!70, fill=orange!70, fill opacity = 0.3] (x1) circle [radius=\epssmall];
    
        \coordinate (x2) at (\radius,-\radius*0.5);
        \fill[blue] (x2) circle (1.5mm) node[anchor=north west, yshift=-4pt, xshift=12pt] {$x_2$};
        \draw[orange!70, fill=orange!70, fill opacity = 0.3] (x2) circle [radius=\epssmall];
    
        \coordinate (x3) at (\radius/2,\bconst);
        \fill[black!30!green] (x3) circle (1.5mm) node[anchor=south, yshift=12pt, xshift=1pt] {$x_3$};
        \draw[orange!70, fill=orange!70, fill opacity = 0.3] (x3) circle [radius=\epssmall];

        \coordinate (x4) at (\radius*1.7,\bconst*1.1);
        \fill[purple] (x4) circle (1.5mm) node[anchor=south, yshift=12pt, xshift=1pt] {$x_4$};
        \draw[orange!70, fill=orange!70, fill opacity = 0.3] (x4) circle [radius=\epssmall];

        
        \begin{scope}[fill opacity=0.1, color=orange, fill=orange!70]
            \filldraw[fill=orange!70] ($1.25*(x1) - 0.25*(x2)$) 
                to[out=240,in=135] ($0.5*(x1) + 0.5*(x2) + (-0.3, -0.8)$)
                to[out=135+180,in=240] ($1.25*(x2) - 0.25*(x1)$)
                to[out=240+180,in=0] ($0.4*(x1) + 0.6*(x2) + (0.5, 0.7)$)
                to[out=180,in=240+180] ($1.25*(x1) - 0.25*(x2)$)
                ;

            \filldraw[fill=orange!70] ($1.3*(x1) - 0.3*(x3)$) 
                to[out=135,in=225] ($0.5*(x1) + 0.5*(x3) + (-0.5, 0.6)$)
                to[out=225+180,in=150] ($1.3*(x3) - 0.3*(x1)$)
                to[out=150+180,in=45] ($0.4*(x1) + 0.6*(x3) + (0.5, -0.7)$)
                to[out=45+180,in=135+180] ($1.3*(x1) - 0.3*(x3)$)
                ;

            \filldraw[fill=orange!70] ($1.27*(x2) - 0.27*(x3)$) 
                to[out=15,in=300] ($0.5*(x2) + 0.5*(x3) + (0.4, 0.7)$)
                to[out=300+180,in=15] ($1.2*(x3) - 0.2*(x2) + (0.1, 0.2)$)
                to[out=15+180,in=100] ($0.4*(x2) + 0.6*(x3) + (-0.5, -0.7)$)
                to[out=100+180,in=15+180] ($1.27*(x2) - 0.27*(x3)$)
                ;

            \filldraw[fill=orange!70] ($1.32*(x4) - 0.32*(x3) + (0.0, 0.0)$) 
                to[out=95,in=15] ($0.5*(x4) + 0.5*(x3) + (0.1, 0.6)$)
                to[out=15+180,in=95] ($1.28*(x3) - 0.28*(x4) + (0.0, 0.0)$)
                to[out=95+180,in=180] ($0.7*(x4) + 0.3*(x3) + (0.1, -0.6)$)
                to[out=180+180,in=95+180] ($1.32*(x4) - 0.32*(x3) + (0.0, 0.0)$)
                ;

            \end{scope}

                \begin{scope}[fill opacity=0.06, color=orange!70]
            \filldraw[fill=orange!70] ($(x1)+(-0.9,-0.9)$) 
                to[out=311,in=225] ($(x2) + (1.0,-1.0)$) 
                to[out=45,in=355] ($(x3) + (0,1.2)$) 
                to[out=180,in=135] ($(x1)+(-0.9,-0.9)$) 
                ;

            \end{scope}

        \path (x1) edge [color=black, dashed, dash pattern=on 2pt off 1pt] (x2);
        \path (x1) edge [color=black, dashed, dash pattern=on 2pt off 1pt] (x3);
        \path (x2) edge [color=black, dashed, dash pattern=on 2pt off 1pt] (x3);
        \path (x4) edge [color=black, dashed, dash pattern=on 2pt off 1pt] (x3);

    \end{tikzpicture}
        \caption{}
        \label{fig: toy example triangle no gap}
    \endminipage\hfill
    \addtocounter{figure}{-1}
    \captionsetup{width=0.95\linewidth,labelformat=empty}
    \caption{Figures 1,2: On the LHS of each figure, a set $S$ of points of different classes with their $\epsilon$-balls in the $\ell_2$ norm. On the RHS, the conflict hypergraph $\confhyp[S]$ represented as orange regions, and the conflict graph $\confgraph[S]$ represented as black dashed lines. For the left figure, the set of hyperedges is $\cal E = \cup_{i \in [5]} \left[ \{i\} \cup \{i, i+1_{\mathrm{mod} 5}\} \right]$, while for the right figure it is $\cal E = 2^{[3]} \cup \{3, 4\} \cup \{4\}$. Best viewed in color.}
\end{figure*}

\begin{restatable}[Adapted from~\cite{dai2024characterizing}]{definition}{defConflictHypergraph} \label{def: conflict hypergraph}
    Let $\mu \in \cal P( \cal X \times \cal Y)$ with support $\suppmu~=~\{(x_i, y_i)\}_{i \in [n]}$. For any $\ell_p$ norm $(p \in (1,\infty])$ and any $\epsilon > 0$, the \textit{conflict hypergraph of $\mu$ at level $\epsilon$} is the hypergraph $\confhyp = (V, \cal E)$ with $V = [n]$ and hyperedge set defined as follows. A set $e \in 2^V$ is a hyperedge of $\confhyp$  if and only if both the following assertions hold: \vspace{-0.1cm}
    \begin{itemize}
        \item[1.] For any distinct $i, j \in e$ we have $y_i \ne y_j,$ 
        \item[2.] $\displaystyle  \bigcap_{i \in e} B_p(x_i,\epsilon) \ne \emptyset$ (\textit{i.e.} the $\epsilon$-balls overlap).
    \end{itemize}
    Similarly, the conflict graph is the graph $\confgraph=(V,E)$ whose edge set is the largest subset of $2^V$ such that for all $e \in E$, both the previous conditions hold and $\vert e \vert =2$.
\end{restatable}

The topology of these graphs and hypergraphs depends only on the support of the distribution $\mu$. Hence, two distributions with the same support will have the same conflict graph and hypergraph. Clearly, the conflict graph\footnote{In graph theory, the conflict graph corresponds to the \emph{2-section} of the conflict hypergraph (see e.g.~\cite{berge1984hypergraphs}).} captures less information about the intrinsic adversarial vulnerability of $\mu$ than the conflict hypergraph. Nevertheless, as we will demonstrate in the subsequent sections, it remains sufficient to identify many of the key structural properties needed for a positive randomization gap to arise.


\textbf{Illustration.} 
We present in Figures \ref{fig: toy example c5} and \ref{fig: toy example triangle no gap} examples of finite discrete distributions and their corresponding conflict graphs and hypergraphs. In both cases, we assume a uniform distribution over the points of the support, all of which belong to different classes. On the LHS of each figure, we depict the support of the distributions and the $\epsilon$-balls using the $\ell_2$ norm. On the RHS of each figure, the dashed lines between the points represent the edges of the conflict graph $\confgraph$, and the orange regions represent the hyperedges of $\confhyp$. 

\begin{remark}
If no two points in the support of $\mu$ are conflicting, then the conflict hypergraph will only contain singletons. In this case, one can easily show that $\mathcal{R}_{\cal F_{\mathrm{det}}}^*(\mu,\epsilon) = \mathcal{R}_{\cal F_{\mathrm{rand}}}^*(\mu,\epsilon)= 0$. See e.g.,~\citep[Appendix E.4]{meunier2021mixed} for a proof when $K=2$. 
\end{remark}

\subsection{Set packing and adversarial risk for deterministic classifiers}

Using the notion of conflict hypergraph for a given distribution $\mu \in \mathcal{P}\left( \mathcal{X} \times \mathcal{Y} \right)$, we now map the adversarial risk minimization problem~\eqref{eq:min-risk-adv-det} in the deterministic setting to a well studied graph theoretical problem known as the \emph{set packing problem}.

\textbf{Set packing problem.} 
Let $H=(V, \cal E)$ be a hypergraph and $\boldsymbol{\omega} \in \mathbb{R}_+^{|V|}$ a weight vector for the vertices. The \emph{set packing problem} over ($H$, $\boldsymbol{\omega}$) seeks a packing (\textit{i.e.}, a subset of vertices that intersects each hyperedge at most once) with maximum cumulative weight. We represent a packing $Q \subseteq V$ using its  characteristic vector $q_Q \in \{0, 1\}^n$, defined component-wise as $q_Q^{(i)} = \1{i \in Q}$ for all $i \in [n]$. Then, the set of possible packings for $H$ can be written as
\begin{equation} \label{eq: set packing}
        \mathcal{Q}(H) = \Big\lbrace q \in \{0, 1\}^n ~|~ \sum_{i \in e}  q^{(i)} \le 1,~ \forall e \in \cal E \Big\rbrace,
\end{equation}
and the set packing problem can be formally put as finding a packing with maximal cumulative weight:\footnote{The set packing problem is traditionally defined with a weight vector $\omegamu = \vecones$ (see e.g.~\cite{schrijver1979fractional}). For simplicity, we use the name \emph{set packing problem} even though we considered the weighted version throughout the paper.}
  \begin{equation} \label{eq: wsp definition}
         \mathrm{IP}(H, \vec{\omega}) := \displaystyle \max_{q \in \mathcal{Q}(H)}  \boldsymbol{\omega}^T q.
     \end{equation}



In Theorem~\ref{thm:adv-risk-minimization-is-FWSP-int} below, we map the optimal adversarial risk for deterministic classifiers on a distribution $\mu \in \mathcal{P}\left( \mathcal{X} \times \mathcal{Y} \right)$ to the value of the set packing problem on the conflict hypergraph of $\mu$.
This result is essentially a byproduct of~\cite[Corollary 1]{dai2024characterizing}. However, our analytical framework significantly differs from~\cite{dai2024characterizing}. Hence, we present a proof for Theorem~\ref{thm:adv-risk-minimization-is-FWSP-int} in Appendix~\ref{app: section 3 computing the randomization gap using graph theory} for the sake of completeness.

\begin{restatable}{theorem}{thmAdvRiskEqualsAdvFWSPInt} \label{thm:adv-risk-minimization-is-FWSP-int}
   Let us consider an $\ell_p$ norm with $p \in (1,\infty]$, and $\epsilon > 0$. For any $\mu \in \mathcal{P}(\cal X \times \mathcal{Y})$ we have:
    \begin{equation*}
        1-\mathcal{R}_{\cal F_{\mathrm{det}}}^*(\mu,\epsilon) = \mathrm{IP}(\confhyp , \omegamu).
    \end{equation*}
\end{restatable}
Theorem~\ref{thm:adv-risk-minimization-is-FWSP-int} allows us to compute the optimal adversarial risk for deterministic classifiers using the set packing problem, which can be expressed as an integer linear problem and for which powerful open-source solvers are available \cite{gurobi}. In simple cases, such as the examples shown in Figures~\ref{fig: toy example c5} and ~\ref{fig: toy example triangle no gap}, the set packing problem can be manually solved, as we present below. 

When considering the packings of $\confhyp$ in Figure~\ref{fig: toy example c5}, we can see that including the point $x_1$ in the packing automatically excludes two points: $x_2$ and $x_5$. This same reasoning applies for all points, which leads to the conclusion that a packing of $\confhyp$ have size at most $2$. 
We conclude that $\mathrm{IP}(\confhyp, \unifvec[5]) = \nicefrac{2}{5}$. Therefore, the optimal adversarial risk for deterministic classifiers over the distribution from Figure~\ref{fig: toy example c5} is $\nicefrac{3}{5}$. For the distribution in Figure~\ref{fig: toy example triangle no gap}, one can easily check that one maximal packing is $\{x_2, x_4\}$. Thus, $\mathrm{IP}(\confhyp, \unifvec[4]) = \nicefrac{2}{4}$ and the optimal (deterministic) adversarial risk is $\nicefrac{1}{2}$. 



\subsection{Fractional set packings and adversarial risk for randomized classifiers}

Similar to the deterministic setting, we can map the adversarial risk minimization problem~\eqref{eq:min-risk-adv-rand} for randomized classifiers to a relaxed version of the set packing problem, called the  \emph{fractional set packing problem}, where the characteristic vector of a packing $Q$ can be fractional, \textit{i.e.}, $q_Q \in [0,1]^n$. Specifically, given a hypergraph $H =(V , \cal E)$ and weights $\boldsymbol{\omega} \in \mathbb{R}_+^{|V|}$, the fractional set packing problem over $(H,\boldsymbol{\omega})$ consists in finding the \textit{fractional packing} with maximum cumulative weight: 
\begin{equation} \label{eq: fwsp definition}
        \displaystyle \mathrm{FP}(H, \boldsymbol{\omega}) \coloneqq \max_{q \in \mathcal{Q}^{\mathrm{frac}}(H)}  \boldsymbol{\omega}^T q,
    \end{equation}
where $\mathcal{Q}^{\mathrm{frac}}(H) = \Big\lbrace q \in [0, 1]^n ~|~ \sum\limits_{i \in e}  q^{(i)} \le 1,~ \forall e \in \cal E \Big\rbrace.$

Then, similar to Theorem~\ref{thm:adv-risk-minimization-is-FWSP-int}, we can show that the optimal adversarial risk for randomized classifiers is related to the value of the fractional set packing problem. The proof of Theorem~\ref{thm:adv-risk-minimization-is-FWSP-frac} is deferred to Appendix~\ref{app: section 3 computing the randomization gap using graph theory}.
\begin{restatable}{theorem}{thmAdvRiskEqualsAdvFWSPFrac} \label{thm:adv-risk-minimization-is-FWSP-frac}
   Let us consider an $\ell_p$ norm with $p \in (1,\infty]$, and $\epsilon > 0$. For any $\mu \in \mathcal{P}(\cal X \times \mathcal{Y})$ we have:
    \begin{equation*}
        1-\mathcal{R}_{\cal F_{\mathrm{rand}}}^*(\mu,\epsilon) = \mathrm{FP}(\confhyp, \omegamu).
    \end{equation*}
\end{restatable}

Theorem~\ref{thm:adv-risk-minimization-is-FWSP-frac} provides a way to compute the optimal adversarial risk for randomized classifiers using linear programming, which can be done efficiently. For the simple examples shown in Figures~\ref{fig: toy example c5} and \ref{fig: toy example triangle no gap}, the fractional set packing can also be solved manually (See Appendix~\ref{app: solving examples fractional set packing}). Hence, the optimal adversarial risks for randomized classifiers for the examples shown in Figures~\ref{fig: toy example c5} and \ref{fig: toy example triangle no gap} are, in both cases, $\nicefrac{1}{2}$. Finally, by combining Theorem~\ref{thm:adv-risk-minimization-is-FWSP-int} and~\ref{thm:adv-risk-minimization-is-FWSP-frac} together, we can rewrite the randomization gap of any finite discrete distribution $\mu \in \mathcal{P}\left( \mathcal{X} \times \mathcal{Y} \right)$ as follows. 




\begin{restatable}{corollary}{corRgReformulation} \label{corollary:rewriting-rg-WSP}
    Let us consider an $\ell_p$ norm with $p \in (1,\infty]$, and $\epsilon > 0$. For any $\mu \in \mathcal{P}(\cal X \times \mathcal{Y})$ we have:
    \begin{equation} \label{eq:rg reformulation}
        \mathrm{rg}(\mu, \epsilon) = \mathrm{FP}(\confhyp, \omegamu) - \mathrm{IP}(\confhyp, \omegamu).
    \end{equation}
\end{restatable}

Since we have already solved both problems for the examples in Figure~\ref{fig: toy example c5} and~\ref{fig: toy example triangle no gap}, we can apply Corollary~\ref{corollary:rewriting-rg-WSP} to compute the randomization gap for these two cases. For Figure~\ref{fig: toy example c5}, we find a positive gap of $\nicefrac{1}{2} - \nicefrac{2}{5} = \nicefrac{1}{10}$, whereas for Figure~\ref{fig: toy example triangle no gap} the randomization gap is 0. The fact that the randomization gap is positive in Figure~\ref{fig: toy example c5} but zero in Figure \ref{fig: toy example triangle no gap}, indicates a fundamental difference between the two distributions. However, it remains unclear how to determine which distributions may have a positive randomization gap. In the next section, we will characterize the structural properties of discrete distributions that exhibit a positive randomization gap. This is made possible by the reformulation in Corollary~\ref{corollary:rewriting-rg-WSP}, which enables us to characterize these distributions by examining their conflict hypergraph representations.



\section{LINK BETWEEN STRUCTURAL PROPERTIES OF THE CONFLICT HYPERGRAPH AND THE RANDOMIZATION GAP
}

 \label{sec: sufficient conditions}









In this section, building upon Corollary~\ref{corollary:rewriting-rg-WSP}, we establish necessary conditions on the conflict hypergraph $\confhyp$ for a distribution $\mu$ to exhibit a positive randomization gap at level $\epsilon$. We will be able to link a positive randomization gap with the presence of specific structures in the conflict hypergraph and the conflict graph of the distribution. Before diving in, let us introduce some graph theoretical terminology that we will use throughout the remainder of the section.

\textbf{Different types of induced subgraphs.
} Let $G=(V,E)$ be an arbitrary graph. An \emph{induced subgraph} of $G$ is a graph $G'=(V', E')$ where $V' \subset V$, $E'= \{ \{i, j \} \in E \mid i,j \in V'\}$. A \textit{clique} of $G$ is an induced subgraph of $G$ in which every two distinct vertices are adjacent, and is said to be maximal if it cannot be extended by adding a vertex from $V$ to $V'$. A \emph{hole} of $G$ is an induced subgraph of $G$ consisting of a cycle of more than three vertices. Accordingly, we can define an \emph{anti-hole} of $G$ as an induced subgraph whose complement is a hole in the complement of $G$. Finally, a graph is considered \emph{perfect} if it contains neither an odd hole nor an odd anti-hole~\cite{chudnovsky2003progress}.



\subsection{Decomposition of the randomization gap using clique hypergraph}

We begin by rewriting the randomization gap as the sum of two non-zero terms. 
Each term in this decomposition will help us uncover key structural features in the conflict hypergraph. To achieve this, we first need to introduce the concept of \emph{clique hypergraph} that plays a central role in the rewriting. 

\begin{definition}
    Let us consider an $\ell_p$ norm with $p \in (1,\infty]$, and $\epsilon > 0$. For any $\mu \in \cal P(\cal X \times \cal Y)$ we define the \textit{clique hypergraph} of $\mu$ at level $\epsilon$ as the hypergraph $\cliquehyp = (V, \cal E)$, where $V =[n]$ and $\cal E$ is the set of all maximal cliques of $\confgraph$.
\end{definition}

Coming back to the examples from Figures~\ref{fig: toy example c5} and~\ref{fig: toy example triangle no gap}, we can illustrate the concept of a clique hypergraph. Specifically, in Figure~\ref{fig: toy example c5}, the clique hypergraph has a hyperedge set \( \{\{1,2\},\{2,3\},\{3,4\},\{4,5\},\{5,1\}\} \), while in Figure~\ref{fig: toy example triangle no gap}, the hyperedge set is \( \{\{1,2,3\}, \{3, 4\}\} \). With this notion of clique hypergraph at hand, we now introduce the technical lemma that allows us to decompose the randomization gap as follows.

\begin{restatable}{lemma}{lemmaInequalitiesFWSP} \label{lemma: reformulations of MIS}
   Let us consider an $\ell_p$ norm with $p \in (1,\infty]$, $\epsilon > 0$, and $\mu \in \mathcal{P}(\cal X \times \mathcal{Y})$.
    Then we have
    \begin{equation} \label{eq: reformulations MIS eq 2}
        \mathrm{IP}(\cliquehyp, \omegamu) = \mathrm{IP}(\confhyp, \omegamu) = \mathrm{IP}(\confgraph, \omegamu).
    \end{equation}
    \begin{equation} \label{eq: reformulations MIS eq 1}
        \mathrm{FP}(\cliquehyp, \omegamu) \le \mathrm{FP}(\confhyp, \omegamu) \le \mathrm{FP}(\confgraph, \omegamu). 
    \end{equation}
\end{restatable}


Combining Corollary~\ref{corollary:rewriting-rg-WSP} and Lemma~\ref{lemma: reformulations of MIS}, we can rewrite the randomization gap using the packing problems on the clique hypergraph $\cliquehyp$ and the conflict hypergraph $\confhyp$ of $\mu$. Then by reintroducing the value of fractional packing problem on the clique hypergraph, we obtain our decomposition of the randomization gap, as follows:
\begin{align}
    \mathrm{rg}(\mu, \epsilon) = & {\color{black}\mathrm{FP}(\confhyp, \omegamu)\textrm{~-~} \mathrm{FP}(\cliquehyp, \omegamu)} \label{eq:randomization-gap-decomposition-eq1} \\ 
    + & {\color{black}\mathrm{FP}(\cliquehyp, \omegamu) \textrm{~-~} \mathrm{IP}(\cliquehyp, \omegamu)} \label{eq:randomization-gap-decomposition-eq2}. 
\end{align} 
The above decomposition holds true as $\mathrm{IP}(\confhyp, \omegamu)$ can be replaced by $\mathrm{IP}(\cliquehyp, \omegamu)$, thanks to \eqref{eq: reformulations MIS eq 2}. 
Furthermore, both~\eqref{eq:randomization-gap-decomposition-eq1} and~\eqref{eq:randomization-gap-decomposition-eq2} are non-negative. Indeed,~\eqref{eq:randomization-gap-decomposition-eq1} is non-negative due to~\eqref{eq: reformulations MIS eq 1} and~\eqref{eq:randomization-gap-decomposition-eq2} is non-negative because the value of the fractional set packing problem is always an upper bound of the set packing problem.
By studying the conditions under which each one of these terms is positive, we can identify the necessary conditions on the conflict hypergraph $\confhyp$ for $\mu$ to exhibit a positive randomization gap.

\subsection{Necessary structures in the distribution for a positive randomization gap}

On the one hand, we note that~\eqref{eq:randomization-gap-decomposition-eq1} becomes zero if the conflict hypergraph $\confhyp$ coincides with the clique hypergraph $\cliquehyp$. Thanks to~\cite{berge1984hypergraphs}, we know that this occurs when every clique in the conflict graph $\confgraph$ is a hyperedge in $\confhyp$; in this case $\confhyp$ is said to be \emph{conformal}~\cite{berge1984hypergraphs}. Hence, for~\eqref{eq:randomization-gap-decomposition-eq1} to be positive, one needs $\confhyp$ to be non-conformal. On the other hand, determining when~\eqref{eq:randomization-gap-decomposition-eq2} is non-zero involves using the characterization of perfect graphs~\cite{chudnovsky2003progress,conforti2014integer}. Specifically,~\eqref{eq:randomization-gap-decomposition-eq2} is zero if the conflict graph $\confgraph$ is perfect. To summarize, we can express conditions for the existence of a distribution with positive randomization gap as follows. 

\begin{restatable}{theorem}{thmSufficientConditions}
    \label{thm: optimal classifier is deterministic}
    Let us consider an $\ell_p$ norm with $p \in (1,\infty]$, and $\epsilon > 0$.
    Let also $S = \{(x_i,y_i)\}_{i \in [n]}$ be an arbitrary set of points from $ \mathcal{X}\times \mathcal{Y}$. There exists a distribution $\mu \in \mathcal{P}(\mathcal{X} \times \mathcal{Y})$ with support $\suppmu=S$ such that $\mathrm{rg}(\mu, \epsilon) > 0$ if and only if at least one of the following assertions holds true:
    
    
    
  \vspace{-0.2cm}
        \begin{enumerate}[\hspace{0.2cm}a)]
            \item $\confhyp[S]$ is not conformal (\textit{i.e.}, there exists a clique in $\confgraph[S]$ that is not a hyperedge in $\confhyp[S]$).
            \item $\confgraph[S]$ is not perfect (\textit{i.e.}, it contains at least one odd hole or one odd anti-hole). 
        \end{enumerate}
   
\end{restatable}

Theorem \ref{thm: optimal classifier is deterministic} is a strong result providing both necessary and sufficient conditions for the \emph{existence} of a distribution $\mu \in \mathcal{P}\left( \mathcal{X} \times \mathcal{Y} \right)$ with positive randomization gap. However, it does not provide a testable condition for a \emph{given} distribution $\mu\ \in \mathcal{P}\left( \mathcal{X} \times \mathcal{Y} \right)$ to determine whether the randomization gap is positive. The following corollary provides (testable) necessary conditions on the support of $\mu$ for a positive randomization gap.


\begin{restatable}{corollary}{CorNecessaryCondition}
    \label{cor:necessary condition}
    Let us consider an $\ell_p$ norm with $p \in (1,\infty]$, and $\epsilon > 0$.
    Let also consider $\mu \in \mathcal{P}\left( \mathcal{X} \times \mathcal{Y} \right)$. If $\mathrm{rg}(\mu, \epsilon) > 0$, then $\confgraph$ satisfies either assertion a) or assertion b) from Theorem~\ref{thm: optimal classifier is deterministic}. 
\end{restatable}


\textbf{Hardness of verifying sufficient conditions.} Given a distribution $\mu$, Corollary~\ref{cor:necessary condition} identifies a necessary condition for the randomization gap to be positive. Furthermore, this condition is checkable in polynomial time~\cite{cornuejols2003polynomial,boros2023dually}. However, we conjecture that any condition that is both \emph{necessary and sufficient} cannot be checked in polynomial time, unless "P=NP"\footnote{It is widely believed that problems solvable in polynomial time (P) are not the same as problems whose solutions are verifiable in polynomial time (NP).}. This conjecture is supported by the fact that, for a given discrete distribution $\mu$ and $\alpha \in \left[0,1\right]$, checking if $\mathrm{rg}(\mu, \epsilon) \ge \alpha$ is co-NP-complete. The proof, based on a reduction from the set packing problem, is in Appendix \ref{app:hardness-section}.


\textbf{Special case of $K=2$.} In binary classification, the conflict graph $\confgraph$ can be shown to be bipartite. Since any bipartite graph is perfect, we have that~\eqref{eq:randomization-gap-decomposition-eq2} is always zero. Additionally, the conflict hypergraph $\confhyp$ is always conformal, as a clique in $\confgraph$ is simply an edge and thus a hyperedge of $\confhyp$.
This implies that~\eqref{eq:randomization-gap-decomposition-eq1} is also zero. Combining both arguments, the contrapositive of the implication stated in Corollary~\ref{cor:necessary condition} tells us that the randomization gap is always zero in the binary case, which validates existing results from~\cite{bhagoji2019lower,gnecco2024role}.

\subsection{The special case of the \texorpdfstring{$\ell_{\infty}$}{infinity} norm}

Interestingly, when considering the $\ell_{\infty}$ norm, we can prove that the only structures that matter in this case are those related to the perfect graph condition. 



\begin{restatable}{corollary}{corLinfNoUncoveredCliques} \label{corollary:linf-no-cliques-uncovered}
    Consider the $\ell_\infty$ norm, and $\epsilon > 0$ and  $\mu \in \mathcal{P}\left( \mathcal{X} \times \mathcal{Y} \right)$ with 
    clique hypergraph $\cliquehyp = (V, \cal E)$. Let $G' = (V', E')$ be an induced subgraph of $\confgraph$. If $G' \text{ is a clique}$, then $V' \in \cal E$. Thus, for any $\boldsymbol \omega \in \mathbb{R}_{+}^{|V|}$:
    \begin{equation*}
        \ \quad \mathrm{FP}(\confhyp, \boldsymbol \omega) = \mathrm{FP}(\cliquehyp, \boldsymbol \omega). 
    \end{equation*}
\end{restatable}

Corollary~\ref{corollary:linf-no-cliques-uncovered} shows that the first source of gap in~\eqref{eq:randomization-gap-decomposition-eq1} is always zero when using the $\ell_\infty$ norm. This implies that the only way for a distribution to exhibit a positive randomization gap is for the conflict graph $\confgraph$ to be non-perfect.
In contrast, when using the $\ell_2$ norm, such uncovered cliques can exist (see Figure~\ref{fig: toy example triangle positive gap}). On the other hand, by Lemma \ref{lemma: build data set from graph G}, both odd holes and anti-holes can exist when using any $\ell_p$ norm for $p \in (1, \infty]$
(see Example~\ref{example:anti-holes-existence}).

\section{THE RANDOMIZATION GAP CAN BE ARBITRARILY CLOSE TO \texorpdfstring{$\nicefrac{1}{2}$}{1/2}}
\label{sec:rg is big}

We now identify a variety of discrete distributions for which we can establish that the randomization gap is arbitrarily close to $\nicefrac{1}{2}$. Specifically, even though the structural conditions introduced in the last section are not sufficient, they provide a systematic way to construct distributions with a significantly large randomization gap, thus generalizing the initial results from~\cite{trillos2023multimarginal,dai2024characterizing}.


\subsection{Large randomization gap based on the conformal condition}

We start by designing a simple data distribution for which the randomization gap is close to $1/2$. Let us consider the discrete distribution $\mu$ that is uniformly distributed over the canonical basis of $\mathbb{R}^K$ and in which each vector is assigned to a distinct class. Specifically, let $(b_1, \dots, b_K)$ denote the canonical basis of $\mathbb{R}^K$ and $\vecones[n]$ denote the vector in $\mathbb{R}^n$ with all components to one. We set $\suppmu = \{(b_1, 1), \dots, (b_K, K)\} \subset \mathbb{R}^K \times [K]$ and $\boldsymbol{\omega}_\mu = \frac{1}{K} \mathbf{1}_K$. Then for $\epsilon = \nicefrac{1}{\sqrt{2}}$ and considering the $\ell_2$ norm, one can verify that the conflict hypergraph $\confhyp$ of $\mu$ contains all 2-hyperedges between every pair of points, with no hyperedges of larger size. In other words, the conflict hypergraph $\confhyp$ is a complete graph over $K$ vertices, including self-loops. Furthermore, every set of vertices forms a clique in the conflict graph $\confgraph$, but none of the cliques of size greater than $2$ are hyperedges in $\confhyp$. Therefore, $\confhyp$ is not conformal (See Figure~\ref{fig: toy example triangle positive gap}). 

\def\scalefig{0.45}
\def\radius{3}
\def\eps{\radius*0.77}
\def\epssmall{\radius*0.16}
\def\aconst{1.73205080757} 
\def\bconst{\radius*\aconst/2}
\def\cyclefactor{0.9}

\begin{figure}[ht!]
        \centering
        \def\radius{3}
        \def\eps{\radius*0.54}
        \def\aconst{1.73205080757} 
        \def\bconst{\radius*\aconst/2}
        
        \def\radiusleft{3.9}
        \def\bconstleft{\radiusleft*\aconst/2}

        \def\radiusleft{\radius}
        \def\bconstleft{\bconst}

        \def\epssmall{\radius/5}
        \begin{tikzpicture}[scale=\scalefig]

        \coordinate (x1) at (0,0);
        \fill[red] (x1) circle (1mm) node[anchor=north east, xshift=2pt, yshift=2pt] {$x_1$};
        \draw[black] (x1) circle [radius=\eps];
    
        \coordinate (x2) at (\radiusleft,0);
        \fill[blue] (x2) circle (1mm) node[anchor=north west, xshift=-2pt, yshift=2pt] {$x_2$};
        \draw[black] (x2) circle [radius=\eps];
    
        \coordinate (x3) at (\radiusleft/2,\bconstleft);
        \fill[black!30!green] (x3) circle (1mm) node[anchor=south] {$x_3$};
        \draw[color=black] (x3) circle [radius=\eps];

    \end{tikzpicture}
    \hspace{-0.5em}
    \begin{tikzpicture}[scale=\scalefig]

        \coordinate (x1) at (0,0);
        \fill[red] (x1) circle (1.5mm) node[anchor=north east, yshift=-6pt, xshift=-4pt] {$x_1$};
        \draw[orange!70, fill=orange!70, fill opacity = 0.3] (x1) circle [radius=\epssmall];
    
        \coordinate (x2) at (\radius,0);
        \fill[blue] (x2) circle (1.5mm) node[anchor=north west, yshift=-4pt, xshift=12pt] {$x_2$};
        \draw[orange!70, fill=orange!70, fill opacity = 0.3] (x2) circle [radius=\epssmall];
    
        \coordinate (x3) at (\radius/2,\bconst);
        \fill[black!30!green] (x3) circle (1.5mm) node[anchor=south, yshift=12pt, xshift=1pt] {$x_3$};
        \draw[orange!70, fill=orange!70, fill opacity = 0.3] (x3) circle [radius=\epssmall];

        
        \begin{scope}[fill opacity=0.1, color=orange, fill=orange!70]
            \filldraw[fill=orange!70] ($(x1)+(-0.5,-0.5)$) 
                to[out=315,in=170] ($(x1) + (0,-0.8)$) 
                to[out=0,in=200] ($(x1) + (1.1,-0.45)$)
                to[out=15,in=150] ($(x1)+(2.5,-0.8)$)
                to[out=320,in=220] ($(x1)+(3.5,-0.7)$)
                to[out=45,in=320] ($(x1)+(3.5,0.8)$)
                to[out=150,in=350] ($(x1)+(1,0.9)$)
                to[out=180,in=30] ($(x1)+(-0.7,0.6)$)
                to[out=215,in=135] ($(x1)+(-0.5,-0.5)$)
                ;
                
            \filldraw[fill=orange!70] ($(x1)+(-0.6,-0.6)$) 
                to[out=315,in=225] ($(x1)+(0.7,-0.6)$)
                to[out=45,in=215] ($(x1)+(1.1,0.9)$)
                to[out=30,in=270] ($(x1)+(2.5,2.5)$)
                to[out=90,in=335] ($(x1)+(1.7,3.5)$)
                to[out=165,in=90] ($(x1)+(0.6,2.5)$)
                to[out=270,in=30] ($(x1)+(-0.7,0.8)$)
                to[out=215,in=135] ($(x1)+(-0.6,-0.6)$)
                ;

            \filldraw[fill=orange!70] ($(x3)+(-0.6,0.4)$) 
                to[out=45,in=155] ($(x3)+(0.7,0.8)$)
                to[out=345,in=90] ($(x3)+(1.5,-0.5)$)
                to[out=270,in=135] ($(x3)+(2,-2)$)
                to[out=315,in=80] ($(x3)+(2.5,-3)$)
                to[out=260,in=15] ($(x3)+(1.8,-3.6)$)
                to[out=190,in=275] ($(x3)+(0.7,-2)$)
                to[out=90,in=330] ($(x3)+(-0.3,-0.7)$)
                to[out=155,in=230] ($(x3)+(-0.6,0.4)$)
                ;
            \end{scope}

        \draw[draw=none] (x1) circle [radius=\eps];
        \draw[draw=none] (x2) circle [radius=\eps];
        \draw[draw=none] (x3) circle [radius=\eps];

        \path (x1) edge [color=black, dashed, dash pattern=on 2pt off 1pt] (x2);
        \path (x1) edge [color=black, dashed, dash pattern=on 2pt off 1pt] (x3);
        \path (x2) edge [color=black, dashed, dash pattern=on 2pt off 1pt] (x3);

    \end{tikzpicture}
      
        \caption{Example of a set of points with a non-conformal conflict hypergraph. The support is the canonical basis of $\mathbb{R}^3$, $\epsilon=\nicefrac{1.1}{\sqrt{2}}$, and $p = 2$. The LHS illustrates the points in the space together with their $\epsilon$-balls, and the RHS shows $\confhyp[s_\mu]$, which is not conformal because it does not contain the hyperedge $\{1, 2, 3\}$.}
        \label{fig: toy example triangle positive gap}
    
\end{figure}
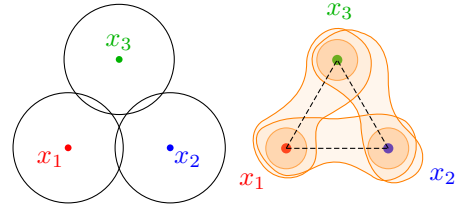

The vector $q = \frac{1}{2} \vecones[K]$ is feasible for the fractional packing problem over $(\confhyp, \boldsymbol{\omega}_\mu)$, \textit{i.e.}, $q \in \mathcal{Q}(\confhyp)$, since the conflict hypergraph $\confhyp$ does not contain any hyperedges of size larger than $2$. This implies that $\mathrm{FP}(\confhyp, \boldsymbol{\omega}_\mu) \ge \boldsymbol{\omega}_\mu^T q = \frac{1}{2}$. Furthermore, $\confhyp$ contains every possible $2$-hyperedge, hence the only possible packings for $\confhyp$ are the singletons. This implies that $\mathrm{IP}(\confhyp, \boldsymbol{\omega}_\mu) = \frac{1}{K}$. Combining these two arguments and using Corollary~\ref{corollary:rewriting-rg-WSP}, the randomization gap of $\mu$ is greater than $\frac{1}{2} - \frac{1}{K}$. Therefore, we can make it arbitrarily close to $\nicefrac{1}{2}$ by choosing $K$ large enough.

The construction presented in this subsection generalizes the initial example from~\cite{trillos2023multimarginal,dai2024characterizing} by scaling it to an arbitrary number of classes \( K \). Nevertheless, it remains restrictive because the distribution \( \mu \) we design is constrained to having one point per class. Additionally, such a construction is not feasible when considering the \( \ell_\infty \) norm, as indicated by Corollary~\ref{corollary:linf-no-cliques-uncovered}. For these reasons, we seek a procedure that enables the construction of less trivial distributions with a large randomization gap, valid for any $\ell_p$-norm.

\subsection{Large randomization gap based on the perfect graph condition}

To design more general distributions, we focus on the perfect graph condition and leverage existing graph constructions from the graph theory literature~\cite{chung1993note}. Doing so, we obtain the following statement. 

\begin{restatable}{theorem}{thmChungDistribution} \label{thm:fibration-arbitrary-rg}
     Fix any $\epsilon>0$ and $\ell_p$ norm with $p \in (1, \infty]$.
    For any $\delta>0$, there exist $d, K \in \mathbb{N}$ and a discrete distribution $\mu \in \cal P(\mathbb{R}^d \times [K])$ such that
    \begin{equation} \label{eq:theorem-ching-chong}
        \mathrm{rg}(\mu, \epsilon) \ge \nicefrac{1}{2} - \delta,
    \end{equation}
    the conflict graph $\confgraph$ is not perfect and the conflict hypergraph $\confhyp$ is conformal.
    Furthermore, the number of classes satisfies $K \in \cal O\left(\sqrt{\nicefrac{n}{\log n}}\right)$ with $n = |\suppmu|$.
\end{restatable}

Theorem~\ref{thm:fibration-arbitrary-rg} show that there exist non-trivial distributions (with $K\neq n$) for which the randomization gap is arbitrarily close to $\nicefrac{1}{2}$. While Theorem~\ref{thm:fibration-arbitrary-rg} itself does not provide an example of such distribution, our proof is constructive and thus provide one. We outline below the main idea of this construction (the complete proof is provided in~\ref{app:section-5-iterative-construction}). 

\textbf{Proof technique.} The main idea of the proof consists in leveraging existing results in graph theory on the construction of large non-perfect, triangle-free graphs. Specifically, we aim to use the iterative procedure presented in~\cite{chung1993note}. Starting from a triangle-free graph $G_0$,~\citet{chung1993note} provides a sequence of triangle-free graphs $\{ G_t \}_{t \in \mathbb{N}}$ such that, for all $t \in \mathbb{N}$,
\begin{equation}\label{eq:rg-gap-chung-graph}
    \mathrm{FP}(G_t, \boldsymbol{\omega}_t) - \mathrm{IP}(G_t, \boldsymbol{\omega}_t) \ge \frac{1}{2} - \left( \frac{2}{3} \right)^t |G_0|,
\end{equation}
where $\boldsymbol{\omega}_t = \unifvec[|G_t|]$ and $|G_0|$ is the size of $G_0$. Note that once the RHS of \eqref{eq:rg-gap-chung-graph} becomes positive, the graphs $G_t$ are guaranteed to be non-perfect. This result is arguably close to what we would like to demonstrate. Intuitively, we would like to design a sequence of distributions $\{ \mu_t \}_{t \in \mathbb{N}}$ such that, for any $t \in \mathbb{N}$, the LHS of~\eqref{eq:rg-gap-chung-graph} can be represented as the randomization gap of $\mu_t$. To design such a sequence, we first show in Lemma~\ref{lemma: build data set from graph G} that for any graph $G_t$, there exists a distribution $\mu_t$ whose conflict graph $\confgraph[s_{\mu_t}]$ is isomorphic to $G_t$. Second, we show that when $G_t$ is triangle-free we can characterize the randomization gap of $\mu_t$ as
\begin{equation} \label{eq:rg-ching-chong}
    \mathrm{rg}(\mu_t, \epsilon) = \mathrm{FP}(G_t, \boldsymbol{\omega}_t) - \mathrm{IP}(G_t, \boldsymbol{\omega}_t).
\end{equation}
This comes from the fact that the conflict graph $\confgraph[s_{\mu_t}]$ is the loopless version of the conflict hypergraph $\confhyp[s_{\mu_t}]$. Therefore, for any $\delta > 0$, one can choose a sufficiently large $t^* \in \mathbb{N}$ such that $ \left( \nicefrac{2}{3} \right)^{t^*} |G_0| < \delta$. Then, combining \eqref{eq:rg-gap-chung-graph} and \eqref{eq:rg-ching-chong}, the randomization gap of $\mu_{t^*} \in \cal P(\mathbb{R}^d \times [K])$ satisfies \eqref{eq:theorem-ching-chong}. Note that the dependency of $K$ on the size of the support $n = | s_{\mu_{t^*}}|$ can be made explicit using the lemma~\ref{lemma: build data set from graph G}.

\begin{figure}
    \centering
    \begin{tikzpicture}[scale = 1.6]
        \begin{scope}[scale=2, every node/.style={fill=white, inner sep=0pt}]
        \foreach \i in {1,2,...,6} {
            \node (Z\i) at ({60*(\i-1)}:1) {};
        }

        \end{scope}

        \begin{scope}[scale=1, every node/.style={circle, draw, fill=white, inner sep=3pt}]
        
        \foreach \i/\j in {4/5, 5/6, 6/1} {
            \node (A\i) at ($(Z\i)!0.20!(Z\j)$) {\scriptsize $a$};
            \node (B\i) at ($(Z\i)!0.50!(Z\j)$) {\scriptsize $b$};
            \node (C\i) at ($(Z\i)!0.80!(Z\j)$) {\scriptsize $c$};
        }
        
        \foreach \i/\j in {1/2, 2/3, 3/4} {
            \node (A\i) at ($(Z\i)!0.20!(Z\j)$) {\scriptsize $a$};
            \node (B\i) at ($(Z\i)!0.50!(Z\j)$) {\scriptsize $b$};
            \node (C\i) at ($(Z\i)!0.80!(Z\j)$) {\scriptsize $c$};
        }

        \end{scope}

        \begin{scope}[color=black]
        \foreach \i/\j in {1/2, 2/3, 3/4} {
            \draw (A\i) -- (B\i) -- (C\i);
            \draw (A\i) to[bend right=60] (C\i);
        }
        \end{scope}

        \begin{scope}[color=black]
        \foreach \i/\j in {4/5, 5/6, 6/1} {
            \draw (A\i) -- (B\i) -- (C\i);
            \draw (A\i) to[bend right=60] (C\i);
        }
        \end{scope}

        \begin{scope}[color=red, dashed]
        \foreach \i/\j in {1/2, 2/3, 3/4, 4/5, 5/6, 6/1} {
            \draw (A\i) to[bend left=30] (B\j);
            \draw (B\i) to[bend left=50] (A\j);
            \draw (B\i) to[bend left=30] (C\j);
            \draw (C\i) to[bend left=50] (B\j);
            \draw (A\i) to[bend left=40] (C\j);
            \draw (C\i) to[bend left=0] (A\j);
        }
        \end{scope}

        \begin{scope}[color=blue]
        \foreach \i/\j in {1/4, 2/5, 3/6} {
            \draw (A\i) to[bend left=0] (A\j);
            \draw (B\i) to[bend left=0] (B\j);
            \draw (C\i) to[bend left=0] (C\j);
        }
        \end{scope}

    \end{tikzpicture}
    \caption{Example of the graph $G_1$ built in \cite{chung1993note} when the initial $G_0$ is the 3-cycle $C_3$. There are 6 copies of $C_3$ with nodes labeled $a,b$ and $c$. Black edges are those within each copy, while \textcolor{red}{red} and \textcolor{blue}{blue} edges are the ones added by the construction in \cite{chung1993note} between different copies of the initial graph.}
    \label{fig:fibration-graph}
\end{figure}

\textbf{Other possible constructions.} To show the existence of non-trivial distributions for which the randomization gap is arbitrarily close to $\nicefrac{1}{2}$, we could have used many other existing explicit constructions~\cite{erdos1966construction, Graham1993ANO, noga1995Ramsey, Alon1994ExplicitRG}, or existence results~\cite{Erdös_1961, Kim1995TheRN} on non-perfect graphs. Those graphs share the property of being triangle-free. This implies they cannot contain any odd anti-hole, and thus their associated conflict hypergraph is conformal. Therefore, the only structures that induce a positive randomization gap are odd holes. We are not aware of any construction in which the graphs contain anti-holes. In remains an open question to determine whether we can design non-trivial distributions for which the randomization gap is arbitrarily close to $\nicefrac{1}{2}$ using anti-hole based graph constructions.




 


\section{DISCUSSIONS \& RELATED WORK}
In this paper, we present a step towards the characterization of data distributions for which randomization proves useful against adversarial examples. Recent work by~\cite{gnecco2024role} demonstrated that in binary classification, the randomization gap (\textit{i.e.}, the difference between the optimal adversarial risk with and without randomization) is always zero for rich enough hypotheses classes. However, they left the multi-class question open. We provide the first (partial) characterization of distributions for which the randomization gap is positive.



\textbf{Closely related work.}
\citet{dai2024characterizing} recently reformulated the optimal adversarial risk as the value of a linear program. However, their objective was to provide lower bounds on this quantity, rather than addressing the difference between randomized and deterministic classifiers. 
They extend prior results from~\citet{bhagoji2019lower} for binary classification, where the authors had already mentioned the equivalence of their formulation to the K\"onig-Egévary theorem in the case of finite spaces. 
\citet{trillos2023multimarginal} discussed in detail the example shown in Figure~\ref{fig: toy example triangle positive gap}, which was also briefly mentioned in \cite{dai2024characterizing}. Our work provides further examples and presents a partial characterization of the distributions for which randomization can enhance robustness.

\textbf{Open problems.} While we provide necessary conditions for the randomization gap to be positive, a full characterization remains to be established. We also leave open the proof of our conjecture that any necessary and sufficient conditions cannot be checked in polynomial time, unless “P = NP”. Although our study focuses on finite discrete distributions, extending the analysis to more general distributions is an interesting future direction. We hypothesize that infinite discrete distributions could be addressed by employing infinite linear programs and infinite graphs. Furthermore, extending the analysis to general Borel probability measures would likely require the application of optimal transport theory~\cite{trillos2023multimarginal}.

\section*{Acknowledgments}
This work was funded by the French National Research Agency (DELCO ANR-19-CE23-0016). This research was supported in part by the French National Research Agency under the France 2030 program, reference ANR-23-PEIA-0003. Rafael is partially supported by the French National Research Agency and the French Ministry of Research and Higher Education. Lucas would like to thank Denis Cornaz, Roland Grappe and Charles Nourry for fruitful discussions.

\bibliographystyle{apalike}
\bibliography{main}

\onecolumn

\newpage
\include{checklist}


\appendix
\newpage

\section{SUPPLEMENTARY MATERIAL FOR SECTION~\ref{sec: computing the randomization gap using graph theory}: COMPUTING THE RANDOMIZATION GAP USING GRAPH THEORY}
\label{app: section 3 computing the randomization gap using graph theory}


We adapt and simplify the proof presented in~\cite{dai2024characterizing} for both theorems. We prove Theorem~\ref{thm:adv-risk-minimization-is-FWSP-frac}, after which Theorem~\ref{thm:adv-risk-minimization-is-FWSP-int} will follow. We first introduce some definitions and establish two technical lemmas. Let us consider an  $\ell_p$ norm with $p \in (1, \infty]$, $\epsilon > 0$, and $\mu \in \mathcal{P}(\mathcal{X} \times \mathcal{Y})$. The adversarial risk of a classifier $f \in \mathcal{F}_\mathrm{rand}$, denoted $\mathcal{R}(\mu, \epsilon, f)$, is defined as 
\begin{equation*}
    \mathcal{R}(\mu, \epsilon, f) \coloneqq \displaystyle \mathbb{E}_{(x, y ) \sim \mu} \left[    \sup_{x' \in B_p(x, \epsilon)}  \mathbb{E}_{z \sim f(x')} \left[ \mathds{1}\{z \neq y \} \right] \right].
\end{equation*}

The \textit{adversarial accuracy} of a classifier $f \in \mathcal{F}_\mathrm{rand}$, denoted $\mathcal{A}(\mu, \epsilon, f)$, is defined as 
\begin{equation*}
    \mathcal{A}(\mu, \epsilon, f) \coloneqq \displaystyle \mathbb{E}_{(x, y ) \sim \mu} \left[    \inf_{x' \in B_p(x, \epsilon)}  \mathbb{E}_{z \sim f(x')} \left[ \mathds{1}\{z = y \} \right] \right] = 1 - \mathcal{R}(\mu, \epsilon, f).
\end{equation*}
and we denote $\mathcal{A}^*_\mathcal{F}(\mu, \epsilon)$ the optimal adversarial accuracy over a family of classifier $\mathcal{F}$. Note that as deterministic classifiers can be represented as randomized ones that only output Dirac measures, $\mathcal{A}(\mu, \epsilon, f)$ and $\mathcal{R}(\mu, \epsilon, f)$ are also well-defined for $f \in \mathcal{F}_\mathrm{det}$.
We thus have that $\mathcal{A}_\mathcal{F}^*(\mu, \epsilon) = 1 - \mathcal{R}_\mathcal{F}^*(\mu, \epsilon)$ for both deterministic and randomized classifiers. Note that for any $\mu \in \mathcal{P}(\mathcal{X} \times \mathcal{Y})$ with support $\suppmu = \{(x_i, y_i)\}_{i \in [n]}$ and probability vector $\omegamu \in \Delta^n$, we can further rewrite the accuracy as follows:
\begin{equation} \label{eq:appendix-accuracy-discrete}
    \mathcal{A}(\mu, \epsilon, f) = \displaystyle \sum_{i \in [n]}  \omegamu^{(i)} \cdot \inf_{x' \in B_p(x_i, \epsilon)} f(x')^{(y_i)}. 
\end{equation}

\paragraph{Reader's note:} In what follows, we make several simplifications to enhance readability.
Let $K$ be the number of classes, \textit{i.e.} $\cal Y = [K]$. We use the natural identification between $\cal P(\cal Y)$ and $\Delta^K$, in which any vector $u \in \Delta^K$ represents a probability distribution over the $K$ classes. Therefore, given a randomized classifier $f: \cal X \to \cal P(\cal Y)$, and a point  $x \in \cal X$, we abuse the notation and call  $f(x)$ the probability vector associated with the point $x$. On the other hand, and recalling that deterministic classifiers can be represented as randomized classifiers that only output Dirac measures over one class, we will often prove results for deterministic classifiers using this characterization.

\subsection{Proofs of Theorem~\ref{thm:adv-risk-minimization-is-FWSP-int} and Theorem~\ref{thm:adv-risk-minimization-is-FWSP-frac}}
We now link the adversarial classification problem with randomized classifiers with the fractional set packing problem. This is done via explicit constructions first introduced in~\cite{dai2024characterizing}, which are presented in the proof of Lemma~\ref{lemma: relation lp and op}. This lemma is the main tool used to prove both Theorem~\ref{thm:adv-risk-minimization-is-FWSP-int} and~\ref{thm:adv-risk-minimization-is-FWSP-frac}.
\begin{restatable}[Extended from~\cite{dai2024characterizing}]{lemma}{lemmaRelationAdvRiskMinAdvFWSP} \label{lemma: relation lp and op}
 Let us consider an $\ell_p$ norm with $p \in (1, \infty]$, and $\epsilon > 0$. Let $\mu \in \cal P (\cal X \times \cal Y)$ with support $s_\mu = \{(x_i, y_i)\}_{i \in [n]}$ and probability vector $\boldsymbol{\omega}_\mu \in \Delta^n$. The following statements hold true:
\begin{enumerate}[\hspace{0.2cm}a)]
    \item For any fractional packing $q \in \mathcal{Q}^{\mathrm{frac}}(\confhyp)$, there exists a classifier $f_q \in \cal F_{\mathrm{rand}}$ such that $\mathcal{A}(\mu, \epsilon, f_q) \ge \omegamu^Tq$.

    \item For any classifier $f \in \cal F_{\mathrm{rand}}$, there exists a fractional packing $q_f \in \mathcal{Q}^{\mathrm{frac}}(\confhyp)$ such that $\mathcal{A}(\mu, \epsilon, f) = \omegamu^Tq_f$.
\end{enumerate}
\end{restatable}
\begin{proof}

Throughout the proof, we denote $\cal E$ the set of hyperedges of the conflict hypergraph $\confhyp$.

\paragraph{Proof of a).} Let $q \in \mathcal{Q}^{\mathrm{frac}}(\confhyp)$ be a fractional packing for \eqref{eq: fwsp definition} over $(\confhyp, \omegamu)$. The first part of the proof will be to define an auxiliary function $g_q$ that will allow us, on a second stage, to define a classifier $f_q$ with the properties that we seek.

\textbf{Definition of the auxiliary function.} Given $(x, y) \in \mathcal{X} \times \mathcal{Y}$, define $\cal I(x,y) = \{i \in [n] ~|~ x \in B_p(x_i, \epsilon) \text{ and } y_i = y\}$ the set of indices of neighbors of $x$ on the set $s_\mu$ that are of class $y$. We denote $\mathcal{Y}_{x} = \{y \in \mathcal{Y} ~|~ \mathcal{I}(x, y) \ne \emptyset\}$ the set of classes for which $x$ has a neighbor in $s_\mu$ that belongs to that class. Denote $g_q : \mathcal{X} \rightarrow \mathbb{R}_{+}^K$ the function defined as follows:
\begin{equation*}
    \forall x \in \cal X, \quad g_q(x)^{(y)} =
    \begin{cases}
      \max_{i \in \cal I(x,y)} q^{(i)}, & \text{if}\  \cal I(x,y) \ne \emptyset.\\
      0, & \text{otherwise.}
    \end{cases}
\end{equation*}
Given that $q$ is a fractional packing of $\confhyp$, we have that for any hyperedge $e \in \cal E$,
\begin{equation} \label{eq:proof lemma link 1}
    \displaystyle \sum_{i \in e} q^{(i)} \le 1.
\end{equation}
Recall that for every $i \in [n]$, $\{i\} \in \cal E$, which by \eqref{eq:proof lemma link 1} implies that 
$$\forall i \in [n], \quad q^{(i)} \le 1.$$
Thus, $0 \le g_q(x)^{(y)} \le 1$ for all $(x, y) \in \cal X \times \cal Y$. Let us now consider a given $x \in \cal X$.  If $x$ is such that $\mathcal{Y}_{x} \ne \emptyset$, we can define for every $y \in \mathcal{Y}_{x}$ the index $i_y$ as 
\begin{equation*}
    i_y = \arg\max_{j \in \mathcal{I}(x, y)} q^{(j)}.
\end{equation*}
Then, for any $x$ such that $\mathcal{Y}_{x} \ne \emptyset$, we can write $g_q(x)$ as
\begin{equation*}
    g_q(x)^{(y)} =
    \begin{cases}
      q^{(i_y)}, & \text{if}\  \cal I(x,y) \ne \emptyset.\\
      0, & \text{otherwise.}
    \end{cases}
\end{equation*}
Furthermore, let $e_x = \{i_y \in [n] ~|~ y \in \mathcal{Y}_{x}\}$. Note that $e_x \in \cal E$, because $x \in \bigcap_{j \in e} B_p(x_j, \epsilon)$.
Given that $q \in \mathcal{Q}^{\mathrm{frac}}(\confhyp)$, the following inequality holds
\begin{equation*}
    \begin{array}{cc}
       \displaystyle \sum_{y \in \mathcal{Y}} g_q(x)^{(y)} = \displaystyle \sum_{y \in \mathcal{Y}_{x}} g_q(x)^{(y)} & = \displaystyle \sum_{i \in e_x}  q^{(i)} \le 1.
    \end{array}
\end{equation*}
\textbf{Definition of the classifier.} We now define a classifier $f_q$ as follows:
\begin{equation*}
    \forall x \in \cal X, \quad f_q(x)^{(y)} =
    \begin{cases}
      1 - \sum_{k>1} g_q(x)^{(k)}& \text{ if } y = 1, \\
      g_q(x)^{(y)}, &  \text{ if }  y > 1.
    \end{cases}
\end{equation*}
By construction, $\sum_{y \in \cal Y} f_q(x)^{(y)} = 1$, and also $f_q(x)^{(y)} \ge g_q(x)^{(y)}$ for any $(x, y) \in \cal X \times \cal Y$ with $\cal Y_x \ne \emptyset$\footnote{Note that the isolated points, \textit{i.e. $\cal Y_x = \emptyset$,} are not involved in the computation of the accuracy. Therefore, the classifier $f_q$  can be defined in any way on those points. For this construction in particular, $f_q$ will always predict class 1 on points $x$ such that $\cal Y_x = \emptyset$.}. The adversarial accuracy of $f_q$ satisfies the following inequality:

\begin{equation*} 
    \begin{array}{clr}
       \mathcal{A}(\mu, \epsilon, f_q) & = \displaystyle \sum_{i \in [n]}  \omegamu^{(i)} \cdot \inf_{x' \in B_p(x_i, \epsilon)} f_q(x')^{(y_i)} & \\
       & \ge \displaystyle \sum_{i \in [n]}  \omegamu^{(i)} \cdot 
 \inf_{x' \in B_p(x_i, \epsilon)} \max_{j \in \cal I(x', y_i)} q^{(j)} & \\
       & \ge \displaystyle \sum_{i \in [n]}  \omegamu^{(i)} \cdot  \inf_{x' \in B_p(x_i, \epsilon)} q^{(i)} & (\text{as } i \in \cal I(x', y_i)) \\
       & \ge \displaystyle \sum_{i \in [n]}  \omegamu^{(i)} \cdot  q^{(i)} & 
    \end{array}
\end{equation*}
This completes the first part of the proof.

\paragraph{Proof of b).} Let us consider an arbitrary classifier $f \in \cal F_{\mathrm{rand}}$, and denote $q_f$ the vector defined as

$$\forall i \in [n], \quad q_f^{(i)} = \inf_{x' \in B_p(x_i, \epsilon)} f(x')^{(y_i)}.$$

Clearly, $q_f^{(i)} \ge 0$ for all $i \in [n]$. 
To see that $q_f$ is feasible for \eqref{eq: fwsp definition} over $(\confhyp, \omegamu)$, consider any hyperedge $e \in \cal E$ and take an arbitrary $x_e \in \bigcap_{i \in e} B_p(x_i, \epsilon)$. We have that  

\begin{equation*}
    \begin{array}{clr}
       \displaystyle \sum_{i \in e} q_f^{(i)} & = \displaystyle \sum_{i \in e} \inf_{x' \in B_p(x_i, \epsilon)} f(x')^{(y_i)} & \\
       & \le \displaystyle \sum_{i \in e} f(x_e)^{(y_i)} & (x_e \in B_p(x_i, \epsilon) \text{ for all } i \in e)\\
       & \le \displaystyle \sum_{y \in \mathcal{Y}} f(x_e)^{(y)}&  (y_i \ne y_j \text{ for any } i, j \in e)\\
       & = 1& ( f \text{ is a valid classifier} )
    \end{array}
\end{equation*}
This holds for any hyperedge in $\cal E$, hence $q_f \in \mathcal{Q}^{\mathrm{frac}}(\confhyp)$. Finally, we can compute the accuracy of $f$:

\begin{equation*} 
       \mathcal{A}(\mu, \epsilon, f)  = \displaystyle \sum_{i \in [n]}  \omegamu^{(i)} \cdot \inf_{x' \in B_p(x_i, \epsilon)} f(x')^{(y_i)}  = \displaystyle \sum_{i \in [n]}  \omegamu^{(i)} \cdot  q_f^{(i)} = \omegamu^T q_f,
\end{equation*}
which concludes the proof.
\end{proof}

\thmAdvRiskEqualsAdvFWSPFrac*
\begin{proof}
    By the first item of Lemma~\ref{lemma: relation lp and op}, we have that for any vector $q \in \cal Q^{\mathrm{frac}}(\confhyp)$, there is a classifier $f_q \in \mathcal{F}_{\mathrm{rand}}$ for which
    \begin{equation*}
        \cal A^*_{ \cal F_{\mathrm{rand}}}(\mu, \epsilon) \ge \cal A(\mu, \epsilon, f_q) \ge \omegamu^T q.
    \end{equation*}
    As this is true for any $q \in \cal Q^{\mathrm{frac}}(\confhyp)$, we have that
    \begin{equation} \label{eq:thm32-eq1}
        \cal A^*_{ \cal F_{\mathrm{rand}}}(\mu, \epsilon) \ge \max_{q \in \cal Q^{\mathrm{frac}}(\confhyp)} \omegamu^T q = \mathrm{FP}(\confhyp, \omegamu).
    \end{equation}
    By the second item of Lemma~\ref{lemma: relation lp and op}, we have that for any classifier $f \in \cal F_{\mathrm{rand}}$, there is a vector $q_f \in \cal Q^{\mathrm{frac}}(\confhyp)$ for which
    \begin{equation*}
        \cal A(\mu, \epsilon, f) = \omegamu^T q_f \le \max_{q \in \cal Q^{\mathrm{frac}}(\confhyp)} \omegamu^T q = \mathrm{FP}(\confhyp, \omegamu).
    \end{equation*}
    As this is  true for any classifier $f \in \cal F_{\mathrm{rand}}$, then by taking $\max$ over all classifiers we have that 
    \begin{equation} \label{eq:thm32-eq2}
        \cal A^*_{ \cal F_{\mathrm{rand}}}(\mu, \epsilon) \le \mathrm{FP}(\confhyp, \omegamu).
    \end{equation}
    
Combining \eqref{eq:thm32-eq1} and \eqref{eq:thm32-eq2} concludes the proof.

\end{proof}

\thmAdvRiskEqualsAdvFWSPInt*
\begin{proof}
    
    The reasoning is analogous to the proof of Theorem~\ref{thm:adv-risk-minimization-is-FWSP-frac} but restricting to deterministic classifiers $\cal F_{\mathrm{det}}$. Let $q \in \cal Q(\confhyp)$ be an arbitrary packing of $\confhyp$. Note that $q \in \cal Q^{\mathrm{frac}}(\confhyp)$, so by Lemma~\ref{lemma: relation lp and op}, there exists a classifier $f_q \in \mathcal{F}_\mathrm{rand}$ such that $\cal A(\mu, \epsilon, f_q) \ge \omegamu^T q$. By the construction of $f_q$ in the proof of Lemma~\ref{lemma: relation lp and op}, we know that $f_q$ is deterministic, \textit{i.e.} $f_q \in \mathcal{F}_\mathrm{det}$. As this is true for any $q \in \cal Q(\confhyp)$, we have that 
    \begin{equation}\label{eq:thm31-eq1}
        \cal A^*_{ \cal F_{\mathrm{det}}}(\mu, \epsilon) \ge \max_{q \in \cal Q(\confhyp)} \omegamu^T q = \mathrm{IP}(\confhyp, \omegamu).
    \end{equation}
    On the other hand, for any deterministic classifier $f \in \mathcal{F}_\textrm{det}$, $f$ can be represented as a randomized classifier, thus by Lemma~\ref{lemma: relation lp and op}, there exists a vector $q_f \in \cal Q^{\mathrm{frac}}(\confhyp)$ for which $\cal A(\mu, \epsilon, f) = \omegamu^T q_f$. Moreover, given the construction of $q_f$ in the proof of Lemma~\ref{lemma: relation lp and op}, we know that $q_f \in \{0,1 \}^n$, which means that $q_f$ is a packing, \textit{i.e.}, $q_f \in \cal Q(\confhyp)$, and therefore $\cal A(\mu, \epsilon, f) \le \mathrm{IP}(\confhyp, \omegamu)$. As this is true for any classifier $f \in \cal F_{\mathrm{det}}$, we have that
    \begin{equation}\label{eq:thm31-eq2}
        \cal A^*_{ \cal F_{\mathrm{det}}}(\mu, \epsilon) \le \mathrm{IP}(\confhyp, \omegamu).
    \end{equation}
    Combining \eqref{eq:thm31-eq1} and \eqref{eq:thm31-eq2} concludes the proof.

\end{proof}

\subsection{Solving Examples of Fractional Set Packing Problems} \label{app: solving examples fractional set packing}

\textbf{Example presented in Figure~\ref{fig: toy example c5}.}
The fractional set packing problem for this case can be written as the following linear program
\begin{equation*}
    \begin{array}{ll@{~~~}ll}
     &\max\limits_{q \in [0, 1]^{5}}  &  \frac{1}{5}\mathbf{1}_5^T q & \\
     &\text{s.t.} & B q \leq \mathbf{1}_{10} &  
    \end{array}
\end{equation*}

where $B \in \{0,1\}^{10 \times 5}$ is the edge-incidence matrix of $\confhyp$. Note that as the largest hyperedges of $\confhyp$ have size 2, every row in $B$ has at most two components equal to 1. This implies that the fractional packing with characteristic vector $q = \frac{1}{2} \mathbf{1}_5$ is feasible, \textit{i.e.}, $q \in \mathcal{Q}^\mathrm{frac}(\confhyp)$, and it has a cumulative weight of $\frac{1}{2}$. Therefore, $\mathrm{FP}(\confhyp, \omegamu)$ is lower bounded by $\frac{1}{2}$. The dual problem of $\mathrm{FP}(\confhyp, \omegamu)$ is the following linear program:

\begin{equation} \label{eq: dual problem example c5}
    \begin{array}{ll@{~~~}ll}
     &\min\limits_{z \in [0, 1]^{10}}  &  \mathbf{1}_{10}^T z & \\
     &\text{s.t.} & B^T z \geq \frac{1}{5} \mathbf{1}_5 &  
    \end{array}
\end{equation}

The vector $z$ can be interpreted as assigning weights to the hyperedges of $\confhyp$, while the vector $B^T z$ represents the total cumulative weight assigned to the vertices of $\confhyp$. Consequently, each constraint in \eqref{eq: dual problem example c5} can be understood as a requirement that each vertex be covered by hyperedges with a cumulative weight of at least $\frac{1}{5}$.

Consider the vector $z^*$, where the components corresponding to the hyperedges associated with singletons are set to $0$, and those corresponding to hyperedges of size $2$ are set to $\frac{1}{10}$. As a result, $z^*$ consists of five components equal to 0 and five components equal to $\frac{1}{10}$.

To verify that $z^*$ is feasible, observe that each vertex is contained in exactly two hyperedges of size 2. Since all these hyperedges have a weight of $\frac{1}{10}$ in $z^*$, we can confirm that every vertex has a total cumulative weight of exactly $\frac{1}{5}$.
The associated value of $z^*$ in \eqref{eq: dual problem example c5} is $\frac{1}{2}$. Therefore, the dual problem is upper bounded by $\frac{1}{2}$, and by duality, we conclude that $\mathrm{FP}(\confhyp, \omegamu) = \frac{1}{2}$.

\textbf{Example presented in Figure~\ref{fig: toy example triangle no gap}.}
Similarly, the fractional set packing problem for this case can be written as a linear program as follows
\begin{equation*} \label{eq: primal problem example triangle}
    \begin{array}{ll@{~~~}ll}
     &\max\limits_{q \in [0, 1]^{4}}  &  \frac{1}{4}\mathbf{1}_4^T q & \\
     &\text{s.t.} & B q \leq \mathbf{1}_{9} &  
    \end{array}
\end{equation*}

where $B \in \{0,1\}^{9 \times 4}$ is the edge-incidence matrix of $\confhyp$. In this case, the fractional packing with characteristic vector $q = (\nicefrac{1}{2}, \nicefrac{1}{2}, 0, 1)$ is feasible, and it has a cumulative weight of $\frac{1}{2}$. Therefore, $\mathrm{FP}(\confhyp, \omegamu)$ is lower bounded by $\frac{1}{2}$. 

The dual problem of $\mathrm{FP}(\confhyp, \omegamu)$ is the following linear program:

\begin{equation} \label{eq: dual problem example triangle}
    \begin{array}{ll@{~~~}ll}
     &\min\limits_{z \in [0, 1]^{9}}  &  \mathbf{1}_{9}^T z & \\
     &\text{s.t.} & B^T z \geq \frac{1}{4} \mathbf{1}_4 &  
    \end{array}
\end{equation}

Consider the vector $z^*$ that is all zeros, except for the component corresponding to the 3-hyperedge $\{1, 2, 3\}$ and the hyperedge $\{4\}$, both of which have value $\frac{1}{4}$.
To verify that $z^*$ is feasible, observe that the 3-hyperedge $\{1, 2, 3\}$ covers these three vertices with a weight of $\frac{1}{4}$. Additionally, the vertex $\{4\}$ is covered by its own hyperedge, so all vertices are properly covered.
The associated value of $z^*$ in \eqref{eq: dual problem example triangle} is $\frac{1}{4} \cdot 2 = \frac{1}{2}$. Therefore, the dual problem is upper bounded by $\frac{1}{2}$, and by duality, we conclude that $\mathrm{FP}(\confhyp, \omegamu) = \frac{1}{2}$.

\section{SUPPLEMENTARY MATERIAL FOR SECTION~\ref{sec: sufficient conditions}: LINK BETWEEN STRUCTURAL PROPERTIES OF THE CONFLICT HYPERGRAPH AND THE RANDOMIZATION GAP} \label{app: sufficient conditions}

The proof of Theorem~\ref{thm: optimal classifier is deterministic} relies on decomposing the randomization gap into the sum of two non-negative terms. In Section~\ref{app:rg-reformulation}, we prove that this decomposition holds true thanks to Lemma~\ref{lemma: reformulations of MIS}. In Section~\ref{app:proof-thm-exist-distribution}, we prove two lemmas adapted from~\cite[Theorem 4.1]{chudnovsky2003progress} that establish equivalences between these terms being zero and the presence of certain structures in the conflict hypergraph. Following this, we present the full proof of Theorem~\ref{thm: optimal classifier is deterministic}. In Section~\ref{app:special-case}, we prove Corollary \ref{corollary:linf-no-cliques-uncovered}, which states that one of the identified structures cannot appear when considering the $\ell_\infty$ norm. Finally, in Section~\ref{app:constructions-examples}, we provide several examples of the existence of specific structures for different $\ell_p$ norms.

\paragraph{Reader's note:} In this section, we will often use the notation $\cal E(H)$ to refer to the set of hyperedges of hypergraph $H$. For readability, we sometimes use the notation $E(G)$ to refer to the edge set of a graph $G$, particularly when handling multiple graphs and hypergraphs simultaneously. Moreover, given a graph \( G = (V, E) \), we often say that \( V' \subset V \) is a clique in \( G \), meaning that the induced subgraph \( G' = (V', E') \), where \( E' = \{ \{i, j \} \in E \mid i, j \in V' \} \), is a clique of \( G \). Finally, given a graph $G = (V,E)$, we define the clique hypergraph of $G$ as $H = (V, \mathcal{E})$, where $\mathcal{E}$ is the set of all maximal cliques of $G$.

\subsection{Proof of Lemma~\ref{lemma: reformulations of MIS}}
\label{app:rg-reformulation}

Lemma~\ref{lemma: appendix order in hypergaphs} is a technical result used in the proof of Lemma~\ref{lemma: reformulations of MIS}. It can be interpreted as establishing a hierarchy between the conflict graph, the conflict hypergraph and the clique hypergraph.

\begin{lemma} \label{lemma: appendix order in hypergaphs}
     Let us consider an $\ell_p$ norm with $p \in (1,\infty]$, and $\epsilon > 0$.
     Let also $S = \{(x_i,y_i)\}_{i \in [n]}$ be an arbitrary set of points from $ \cal X \times \cal Y$.
     The following assertion hold true
    \begin{enumerate}
    [\hspace{0.2cm}a)]
        \item For any edge $e \in  E(\confgraph[S])$, there exists some hyperedge $e' \in \cal E(\confhyp[S])$ such that $e \subseteq e'$.
        \item For any hyperedge $e \in \cal E(\confhyp[S])$, there exists some hyperedge $e' \in \cal E(\cliquehyp[S])$ such that $e \subseteq e'$.
    \end{enumerate}
\end{lemma}
\begin{proof}
    \textbf{Proof of a).}
    This is true by definition of conflict hypergraph and conflict graph. In particular, by the fact that $\confgraph[S]$ is the 2-section of $\confhyp[S]$.

    \textbf{Proof of b).} Now consider an arbitrary hyperedge $e \in \cal E (\confhyp[S])$. By definition of conflict graph, this implies that
    \begin{equation*}
        \forall i, j \in e \text{ with } i \ne j, \quad \{i, j\} \in E(\confgraph[S]).
    \end{equation*}
    The fact that all the pairs $i, j \in e$ are edges of $\confgraph[S]$ means that $e$ constitutes a clique of $\confgraph[S]$. Then, by definition of the clique hypergraph as the one whose hyperedges are the maximal cliques of $\confgraph[S]$, we have that 
    \begin{equation*}
        \exists e' \in \cal E(\cliquehyp[S]) \text{ such that } e \subseteq e'.
    \end{equation*}
\end{proof}

The hierarchy of the three hypergraphs, as established in Lemma \ref{lemma: appendix order in hypergaphs}, corresponds to an ordering of the values in their respective fractional set packing problems. According to Lemma \ref{lemma: reformulations of MIS}, the three hypergraphs represent different formulations of the same set packing problem, with their fractional versions providing upper bounds of varying tightness.

\lemmaInequalitiesFWSP*
\begin{proof}
    \textbf{Proof of \eqref{eq: reformulations MIS eq 1}}.
    We will begin by using Lemma~\ref{lemma: appendix order in hypergaphs} to show that the set of fractional packings of the hypergraphs $\cliquehyp$, $\confhyp$ and $\confgraph$ satisfy the following relation:
    \begin{equation*}
        \cal Q^{\mathrm{frac}}(\cliquehyp) \subseteq \cal Q^{\mathrm{frac}}(\confhyp) \subseteq \cal Q^{\mathrm{frac}}(\confgraph).
    \end{equation*}
    This will prove that \eqref{eq: reformulations MIS eq 1} holds, as the fractional packing problem is a maximization problem. 
    
    ${( \cal Q^{\mathrm{frac}}(\cliquehyp) \subseteq \cal Q^{\mathrm{frac}}(\confhyp))}.$ Take an arbitrary $q \in \cal Q^{\mathrm{frac}}(\cliquehyp)$ and an arbitrary hyperedge $e \in \cal E(\confhyp)$. By Lemma~\ref{lemma: appendix order in hypergaphs}, there exists a hyperedge $e' \in \cal E(\cliquehyp)$ such that $e \subseteq e'$. We then have that 
    \begin{equation*}
        \displaystyle \sum_{i \in e} q^{(i)} \le  \sum_{i \in e'} q^{(i)} \le 1,
    \end{equation*}
    where the last inequality holds because $q \in \cal Q^{\mathrm{frac}}(\cliquehyp)$ and $e' \in \cal E(\cliquehyp)$. As this holds for every hyperedge $e \in \cal E(H)$, we conclude that $q$ is a fractional packing of $\confhyp$, \textit{i.e.} $q \in \cal Q^{\mathrm{frac}}(\confhyp)$.
    
    ${( \cal Q^{\mathrm{frac}}(\confhyp) \subseteq \cal Q^{\mathrm{frac}}(\confgraph))}.$ Similarly, take an arbitrary $q \in \cal Q^{\mathrm{frac}}(\confhyp)$ and an arbitrary edge $e \in \cal E(\confgraph)$. By Lemma~\ref{lemma: appendix order in hypergaphs}, there exists a hyperedge $e' \in \cal E(\confhyp)$ such that $e \subseteq e'$. We then have that 
    \begin{equation*}
        \displaystyle \sum_{i \in e} q^{(i)} \le  \sum_{i \in e'} q^{(i)} \le 1,
    \end{equation*}
    where the last inequality holds because $q \in \cal Q^{\mathrm{frac}}(\confhyp)$ and $e' \in \cal E(\confhyp)$. As this holds for every hyperedge $e \in \cal E(\confgraph)$, we conclude that $q$ is a fractional packing of $\confgraph$, \textit{i.e.} $q \in \cal Q^{\mathrm{frac}}(\confgraph)$. We thus have that $\cal Q^{\mathrm{frac}}(\cliquehyp) \subseteq \cal Q^{\mathrm{frac}}(\confhyp) \subseteq \cal Q^{\mathrm{frac}}(\confgraph)$, which proves that \eqref{eq: reformulations MIS eq 1} holds.

    \textbf{Proof of \eqref{eq: reformulations MIS eq 2}.} We will prove that 
    \begin{equation*}
        \cal Q(\cliquehyp) = \cal Q(\confhyp) = \cal Q(\confgraph).
    \end{equation*}
    Note that the reasoning used to deal with \eqref{eq: reformulations MIS eq 1} can be directly applied to prove that
    \begin{equation*}
        \cal Q(\cliquehyp) \subseteq \cal Q(\confhyp) \subseteq \cal Q(\confgraph).
    \end{equation*}
    
    Therefore, it suffices to show that $\cal Q(\confgraph) \subseteq \cal Q(\cliquehyp)$. Suppose by contradiction that this is not the case. Then there must exists some binary vector $q \in \{0, 1\}^n$ such that $q \in \cal Q(\confgraph)$ but $q \notin \cal Q(\cliquehyp)$. In particular, the condition $q \notin \cal Q(\cliquehyp)$ implies that
    \begin{equation*}
        \exists e \in \cal E(\cliquehyp) \text{ such that } \sum_{i \in e} q^{(i)} > 1. 
    \end{equation*}
    Given that $q$ is a binary vector, this implies that 
    \begin{equation*}
        \exists i, j \in e,~i \ne j  \text{ such that }  q^{(i)} = q^{(j)} = 1.
    \end{equation*}
    These two indices $i, j$ form an edge in $\confgraph$, which in turn implies that $q$ cannot be a packing of $\confgraph$.
    This contradiction allows us to conclude that $\cal Q(\confgraph) \subseteq \cal Q(\cliquehyp)$, and therefore
    \begin{equation*}
        \cal Q(\cliquehyp) = \cal Q(\confhyp) = \cal Q(\confgraph).
    \end{equation*}
    This concludes the proof.
\end{proof}

\subsection{Proof of Theorem \ref{thm: optimal classifier is deterministic}.}
\label{app:proof-thm-exist-distribution}

Lemmas~\ref{lemma: rg decomposition - clique} and~\ref{lemma: rg decomposition - perfect} are adapted from \cite[Theorem 4.1]{chudnovsky2003progress} and are used to prove Theorem \ref{thm: optimal classifier is deterministic}. Each lemma presents an equivalence between the positivity of \eqref{eq:randomization-gap-decomposition-eq1} or \eqref{eq:randomization-gap-decomposition-eq2}, and the existence of a particular structure in the conflict hypergraph.

\begin{lemma}[Conformal hypergraphs] \label{lemma: rg decomposition - clique}
    Let us consider an $\ell_p$ norm with $p \in (1,\infty]$, and $\epsilon > 0$.
    Let also $S = \{(x_i,y_i)\}_{i \in [n]}$ be an arbitrary set of points from $ \cal X \times \cal Y$. The assertions a) and b) below are equivalent. 

    \begin{enumerate}[\hspace{0.2cm}a)]
        \item  $\forall \boldsymbol{\omega} \in \mathbb{R}_+^n, \quad \mathrm{FP}(\confhyp[S], \boldsymbol{\omega}) = \mathrm{FP}(\cliquehyp[S], \boldsymbol{\omega})$
        \item Every clique in $\confgraph[S]$ is a hyperedge in $\confhyp[S]$.
    \end{enumerate}
    
\end{lemma}
\begin{proof}
        ${\bf b) \implies a).}$ Let $\boldsymbol{\omega} \in \mathbb{R}_+^n$. From Lemma \ref{lemma: reformulations of MIS}, we have that  $\mathrm{FP}(\confhyp[S], \boldsymbol{\omega}) \geq \mathrm{FP}(\cliquehyp[S], \boldsymbol{\omega})$. Furthermore, for any fractional packing of $\confhyp[S]$ with characteristic vector $q \in \mathcal{Q}^\mathrm{frac}(\confhyp[S])$, assertion \textbf{b)} implies that $q \in \mathcal{Q}^\mathrm{frac}(\cliquehyp[S])$. Thus, $\mathcal{Q}^\mathrm{frac}(\confhyp[S]) \subseteq \mathcal{Q}^\mathrm{frac}(\cliquehyp[S])$ and then $\mathrm{FP}(\confhyp[S], \boldsymbol{\omega}) \leq \mathrm{FP}(\cliquehyp[S], \boldsymbol{\omega})$ which conclude this first implication. 
    
    ${\bf a) \implies b).}$
    By contradiction, suppose there exists a clique $c'$ in $\confgraph[S]$ such that $c'$ is not a hyperedge in $\confhyp[S]$ (\textit{i.e.} $c' \notin \mathcal{E}(\confhyp[S])$). Let $c$ be the maximal clique in $\confgraph[S]$ containing $c'$ (\textit{i.e}. $c \in \mathcal{E}(\cliquehyp[S])$). Note that, as $\confhyp[S]$ is downward closed, we have that $c \notin \cal E(\confhyp[S])$. Consider the probability vector $\boldsymbol{\omega} \in \Delta^n$ where $\boldsymbol{\omega}^{(i)} = \frac{1}{|c|}\1{i \in c}$ for all $i \in [n]$. Note that $\boldsymbol{\omega}$ is an optimal solution for the fractional set packing problem over $(\cliquehyp[S], \boldsymbol \omega)$ as $\boldsymbol{\omega}^T \boldsymbol{\omega} = \frac{1}{|c|}$ and for any $q \in \mathcal{Q}^\mathrm{frac}(\cliquehyp[S])$ we have
    \begin{equation*}
        \boldsymbol{\omega}^T q = \frac{1}{|c|} \sum_{i \in c} q^{(i)} \leq \frac{1}{|c|} \quad (\text{as } c  \in \mathcal{E}(\cliquehyp[S])).
    \end{equation*}
    We now construct a feasible solution $q$ for the fractional set packing problem over $(\confhyp[S], \boldsymbol{\omega})$ such that $\boldsymbol{\omega}^T q > \boldsymbol{\omega}^T \boldsymbol{\omega}$, which would imply that $\mathrm{FP}(\confhyp[S], \boldsymbol{\omega}) \ge \boldsymbol{\omega}^T q > \boldsymbol{\omega}^T \boldsymbol{\omega} = \mathrm{FP}(\cliquehyp[S], \boldsymbol{\omega})$, hence contradicting \textbf{a)}. Let $\mathcal{E}_c := \{e \in \mathcal{E}(\confhyp[S]) : e \subseteq c  \}$ and consider $e^* \in \argmax \{ |e| : e \in \mathcal{E}_c\}$. Note that $e^* \subset c$ as $c \notin \mathcal{E}_c$, which implies 
    \begin{equation}
    \label{eq:proof_conformal_hypergraph}
        \sum_{i \in e^*} \boldsymbol{\omega}^{(i)} < \sum_{i \in c} \boldsymbol{\omega}^{(i)} = 1.
    \end{equation}
    Let $i^* \in e^*$ and define $q \in [0,1]^n$ such that $q^{(i)} = \boldsymbol{\omega}^{(i)}$ for $i \neq i^*$ and $q_{i^*} = 1 - \sum_{i \in e^* \setminus \{ i^* \}} \boldsymbol{\omega}^{(i)}$. Notice that 
    \begin{align*}
        \boldsymbol{\omega}^T q = \frac{1}{|c|} \sum_{i \in c} q^{(i)} &=  \frac{1}{|c|} \left(\sum_{i \in c\setminus e^*} q^{(i)} +   \sum_{i \in e^*} q^{(i)}   \right)\\
        &= \frac{1}{|c|} \left(\sum_{i \in c\setminus e^*} \boldsymbol{\omega}^{(i)} +   1  \right)\\
        &> \frac{1}{|c|} \left(\sum_{i \in c\setminus e^*} \boldsymbol{\omega}^{(i)} +   \sum_{i \in e^*} \boldsymbol{\omega}^{(i)}  \right) \quad \text{(by \eqref{eq:proof_conformal_hypergraph})}\\
        &= \boldsymbol{\omega}^T \boldsymbol{\omega}.
    \end{align*}
    In order to conclude the proof, we need to check that $q$ is feasible, \textit{i.e.} $q \in \mathcal{Q}^\textrm{frac}(\confhyp[S])$. Let us consider an arbitrary $e \in \mathcal{E}(\confhyp[S])$. Note that the following is true given the definition of $q$:
        \begin{equation*}
        \sum_{i \in e} q^{(i)} = \sum_{i \in e \cap c} q^{(i)}. 
    \end{equation*}
    Now let us consider two cases, depending on whether $i^*$ belongs to $e$ or not.
    If $i^* \in e \cap c $, then
    \begin{align*}
        \sum_{i \in e \cap c} q^{(i)} &= 1 - \sum_{i \in e^* \setminus  \{ i^* \}} q^{(i)} +  \sum_{\substack{i \in e \cap c  \\ i \neq i^*}} q^{(i)} & \nonumber \\
        & = 1 - \frac{|e^*| - 1}{|c|} +  \frac{|e \cap c| - 1}{|c|} &  \nonumber 
 \\ 
        &= 1 - \frac{|e^*| - |e \cap c|}{|c|} & \nonumber \\
        & \leq 1. & (\text{By definition of }e^*)
    \end{align*}
Otherwise, if $i^* \notin e \cap c$, then 
\begin{equation*}
    \sum_{i \in e \cap c} q^{(i)} =\frac{|e \cap c|}{|c|} \leq 1.
\end{equation*}
This concludes the proof.
\end{proof}

\begin{lemma}[Perfect graphs] \label{lemma: rg decomposition - perfect}
    Let us consider an $\ell_p$ norm with $p \in (1,\infty]$, and $\epsilon > 0$.
    Let also $S = \{(x_i,y_i)\}_{i \in [n]}$ be an arbitrary set of points from $ \mathcal{X}\times \mathcal{Y}$. The assertions a) and b) below are equivalent. 

    \begin{enumerate}[\hspace{0.2cm}a)]
    \item $\forall~ \boldsymbol{\omega} \in \mathbb{R}_+^n, \quad \mathrm{FP}(\cliquehyp[S], \boldsymbol{\omega}) = \mathrm{IP}(\cliquehyp[S], \boldsymbol{\omega}).$ 
    
    \item $\confgraph[S]$ is perfect. 
    \end{enumerate}
        
\end{lemma}
\begin{proof}
    Let $A$ be the $m \times n$ edge-incidence matrix of the hypergraph $\cliquehyp[S]$. Recall the linear program formulation of the fractional set packing problem
    \begin{equation} \label{eq: fp linear program proof perfect lemma}
        \begin{array}{ll@{~~~}ll}
         \mathrm{FP}(\cliquehyp[S], \boldsymbol{\omega}) = &\max\limits_{q \in [0, 1]^{n}}  &  \boldsymbol{\omega}^T q & \\
         &\text{s.t.} & A q \leq \vecones[m] &  
        \end{array}
    \end{equation}
    By \cite[Theorem 4.1]{chudnovsky2003progress}, the linear program in \eqref{eq: fp linear program proof perfect lemma} has an integral optimum solution \emph{for every} objective function $\boldsymbol{\omega} \in \mathbb{R}_+^n$ if and only if the matrix $A$ is the edge-incidence matrix of the clique hypergraph of a perfect graph.
    
    As we already assumed that $A$ is the incidence matrix of $\cliquehyp[S]$, which is the clique hypergraph of $\confgraph[S]$, then the only condition to ensure the existence of an integral optimum solution \emph{for every} objective function $\boldsymbol{\omega} \in \mathbb{R}_+^n$ for $ \mathrm{FP}(\cliquehyp[S], \boldsymbol{\omega})$ is that $\confgraph[S]$ is perfect.

\end{proof}

\thmSufficientConditions*
\begin{proof}

    \textbf{First implication.}
    Let $\mu \in \cal P (\mathcal{X}\times \mathcal{Y})$ be the distribution with support $\suppmu$ and probability vector $\omegamu$ such that $\mathrm{rg}(\mu, \epsilon) > 0$. By the decomposition of the randomization gap, one of either \eqref{eq:randomization-gap-decomposition-eq1} or \eqref{eq:randomization-gap-decomposition-eq2} has to be positive.

    If \eqref{eq:randomization-gap-decomposition-eq1} is positive, \textit{i.e.} ${\color{black!20!orange}\mathrm{FP}(\confhyp, \omegamu)\textrm{~-~} \mathrm{FP}(\cliquehyp, \omegamu)} > 0$, then by Lemma \ref{lemma: rg decomposition - clique}, there exists a clique in $\confgraph$ that is not a hyperedge of $\confhyp$. This means that $\confhyp$ is not conformal.

    If \eqref{eq:randomization-gap-decomposition-eq2} is positive, \textit{i.e.} ${\color{purple}\mathrm{FP}(\cliquehyp, \omegamu) \textrm{~-~} \mathrm{IP}(\cliquehyp, \omegamu)} > 0$, then by Lemma \ref{lemma: rg decomposition - perfect}, the graph $\confgraph$ is not perfect, which means that it has either an odd hole or an odd anti-hole.


    \textbf{Second implication. }
    If \emph{a)} holds and there is a clique of $\confgraph[S]$ that is not a hyperedge of $\confhyp[S]$, then by Lemma \ref{lemma: rg decomposition - clique} there must exist some vector $\boldsymbol{\omega}$ such that $\mathrm{FP}(\confhyp[S], \boldsymbol{\omega}) \neq \mathrm{FP}(\cliquehyp[S], \boldsymbol{\omega})$. By Lemma \ref{lemma: reformulations of MIS}, it is always true that $\mathrm{FP}(\confhyp[S], \boldsymbol{\omega}) \geq \mathrm{FP}(\cliquehyp[S], \boldsymbol{\omega})$, which implies that ${\color{black!20!orange}\mathrm{FP}(\confhyp[S], \boldsymbol{\omega}) \textrm{~-~} \mathrm{FP}(\cliquehyp[S], \boldsymbol{\omega})} > 0.$

    If \emph{b)} holds and $\confgraph[S]$ is not perfect, by Lemma \ref{lemma: rg decomposition - perfect}, there must exist some vector $\boldsymbol{\omega}$ such that $\mathrm{FP}(\cliquehyp[S], \boldsymbol{\omega}) \neq \mathrm{IP}(\cliquehyp[S], \boldsymbol{\omega})$. Given that $\mathrm{FP}(\cliquehyp[S], \boldsymbol{\omega}) \geq \mathrm{IP}(\cliquehyp[S], \boldsymbol{\omega})$, we have that ${\color{purple} \mathrm{FP}(\cliquehyp[S], \boldsymbol{\omega})\textrm{~-~}  \mathrm{IP}(\cliquehyp[S], \boldsymbol{\omega})} >0$.

   Define $\mu$ as the distribution with support on $S$ and probability vector $\boldsymbol{\omega}$ such that either ${\color{black!20!orange}\mathrm{FP}(\confhyp[S], \boldsymbol{\omega}) \textrm{~-~} \mathrm{FP}(\cliquehyp[S], \boldsymbol{\omega})} > 0$ or ${\color{purple} \mathrm{FP}(\cliquehyp[S], \boldsymbol{\omega})\textrm{~-~}  \mathrm{IP}(\cliquehyp[S], \boldsymbol{\omega})} >0$. Then, we have that:
    \begin{equation*}
        \mathrm{rg}(\mu, \epsilon) = {\color{purple}\mathrm{FP}(\cliquehyp[S], \boldsymbol{\omega}) \textrm{~-~} \mathrm{IP}(\cliquehyp[S], \boldsymbol{\omega})} + {\color{black!20!orange}\mathrm{FP}(\confhyp[S], \boldsymbol{\omega}) \textrm{~-~} \mathrm{FP}(\cliquehyp[S], \boldsymbol{\omega})} > 0.
    \end{equation*}
  
\end{proof}

\subsection{The special case of the \texorpdfstring{$\ell_{\infty}$}{infinity} norm.}
\label{app:special-case}

Having identified that the randomization gap is linked with the presence of three possible problematic structures in the conflict hypergraph, it is worth asking if these structures actually exist. It turns out that when considering the $\ell_{\infty}$ norm, the uncovered cliques related to the nonconformity of the conflict hypergraph can never occur. 

\corLinfNoUncoveredCliques*
\begin{proof}
    Let $\{(x_i, y_i)\}_{i \in [n]} \subset \mathcal{X} \times \mathcal{Y}$ be the support of $\mu$ such that $1, \dots, n'$ are the indices in $V'$ with $n' = |V'|$. Suppose that $G'$ is a clique.  Recall that $\mathcal{X} \subseteq \mathbb{R}^d$. For any $j \in [d]$, consider a permutation $\sigma_j : [n'] \rightarrow [n']$ that satisfies $x_{\sigma_j(1)}^{(j)} \leq \dots \leq x_{{\sigma_j(n')}}^{(j)}$. Then, given that $\{\sigma_j(1), \sigma_j(n')\} \in E'$, we have that $B_\infty(x_{\sigma_j(1)}, \epsilon) \cap B_\infty(x_{{\sigma_j(n')}}, \epsilon) \ne \emptyset$, which implies
    \begin{equation*}
         I_j \coloneqq [x_{{\sigma_j(n')}}^{(j)} - \epsilon, x_{\sigma_j(1)}^{(j)} + \epsilon] \neq \emptyset, \quad \text{ for any } j \in [d].
    \end{equation*}
    Let \( z \in \mathbb{R}^d \) be such that \( z^{(j)} \in I_j \) for every $j \in [d]$. Given that for any $j \in [d]$, \( z^{(j)} \) satisfies the inequality \( x_{{\sigma_j(n')}}^{(j)} - \epsilon \leq z^{(j)} \leq x_{\sigma_j(1)}^{(j)} + \epsilon \) and that \( x_{\sigma_j(1)}^{(j)} \leq \dots \leq x_{{\sigma_j(n')}}^{(j)} \), we can conclude that:
    \begin{equation*}
        \forall i \in V', \quad \forall j \in [d], \quad x_{i}^{(j)} - \epsilon \leq z^{(j)} \leq x_{i}^{(j)} + \epsilon.
    \end{equation*}
    In other words, we have that $z \in B_\infty(x_i, \epsilon)$ for all $i \in V'$. This implies that the intersection of the $\epsilon$-balls is non-empty, \textit{i.e.},
        $\bigcap_{i \in V'} B_\infty(x_i, \epsilon) \neq \emptyset.$
    Since $G'$ is a clique of the conflict graph, for any \( i,j \in V' \) we have that $\{i, j\} \in E'$ and thus \( y_i \neq y_j \). Therefore, \( V' \) is a hyperedge in \( \confhyp\).
\end{proof}

\subsection{Constructions of specific structures for different \texorpdfstring{$\ell_{p}$}{p} norms} \label{app:constructions-examples}

\begin{example}[Non-conformal conflict hypergraph in the $\ell_2$ norm.] \label{example: uncovered cliques l2}
Let us consider the $\ell_2$ norm. Consider $\mu \in \mathcal{P}(\mathbb{R}^K \times [K])$ the uniform distribution over the canonical basis of $\mathbb{R}^K$, \textit{i.e.}, $\suppmu = \{(b_1, 1), \dots, (b_K, K)\} \subset \mathbb{R}^K \times [K]$ and $\boldsymbol{\omega}_\mu = \frac{1}{K} \mathbf{1}_K$. 
\paragraph{Intersection of $\epsilon$-balls for subsets of points.}    
For any subset $S \subseteq [K]$ such that $|S| = m$, we denote by $b_S$ the average of the canonical vectors indexed by $S$, \textit{i.e.},
\begin{equation*}
    b_S \coloneqq \frac{1}{m} \sum_{i \in S} b_i.
\end{equation*}
Furthermore, we can prove the following equivalence (see below):
\begin{equation} \label{eq:example-uncovered-clique-l2-barycenter-condition}
    \bigcap_{i \in S} B_2(b_i, \epsilon) \ne \emptyset \iff b_S \in \bigcap_{i \in S} B_2(b_i, \epsilon).
\end{equation}

Simple calculations tell us that, for any $S$ of size $m$ and $i \in [n]$, the Euclidean distance between $b_S$ and $b_i$ is exactly $\sqrt{\frac{m-1}{m}}$. Hence, using~\eqref{eq:example-uncovered-clique-l2-barycenter-condition}, we have
\begin{equation} \label{eq:example-uncovered-clique-l2-m}
    \bigcap_{i \in S} B_2(b_i, \epsilon) \ne \emptyset \iff \epsilon \ge \sqrt{\frac{m-1}{m}}.
\end{equation}
\paragraph{The conflict hypergraph is not conformal.}
Let us set $\epsilon = \frac{1}{\sqrt{2}}$. We have that
\begin{equation*}
    \forall i, j \in [K], \quad \lVert b_i - b_j \rVert_2 = 2\epsilon,
\end{equation*}
and therefore $\{i, j\} \in \cal E(\confhyp)$ for all $i, j \in [K]$. This means that the subset of vertices $[K]$ induces a clique in $\confgraph$.

On the other hand, for any subset $S \subset [K]$ of size $m$ greater than 2, we have that $\sqrt{\frac{m-1}{m}} > \epsilon$, so by \eqref{eq:example-uncovered-clique-l2-m} we have that 
\begin{equation*}
    \forall S \subset [K], \quad |S| > 2 \implies \bigcap_{i \in S} B_2(b_i, \epsilon) = \emptyset,
\end{equation*}
which implies that no subset of size greater than 2 can be a hyperedge of $\confhyp$. In particular, the subset $[K]$, which constitutes a clique in $\confgraph$, is not a hyperedge of $\confhyp$. Thus, $\confhyp$ is not conformal.

\paragraph{Proof of \eqref{eq:example-uncovered-clique-l2-barycenter-condition}.} Now we proceed to prove the result stated in \eqref{eq:example-uncovered-clique-l2-barycenter-condition}. We will do so by contradicting the fact that the mean minimizes the sum of squared distances.
One direction is obvious, so we are going to prove that if the intersection $\cap_{i \in S} B_2(b_i, \epsilon)$ is non-empty, then the average $b_S$ must belong to it.

Suppose by contradiction that $b_S \notin \cap_{i \in S} B_2(b_i, \epsilon)$ but that $\cap_{i \in S} B_2(b_i, \epsilon) \ne \emptyset$. Then, there exists an index $i^* \in S$ such that $b_S \notin B_2(b_{i^*}, \epsilon)$. This implies that 
\begin{equation} \label{eq:example-uncovered-clique-l2-1}
    \lVert b_S - b_{i^*} \rVert_2 > \epsilon.
\end{equation}
Note that the distance from $b_S$ to any $b_i$, $i \in S$ is exactly $\sqrt{\frac{m-1}{m}}$, so \eqref{eq:example-uncovered-clique-l2-1} implies the following:
\begin{equation} \label{eq:example-uncovered-clique-l2-2}
    \forall i \in S, \quad \lVert b_S - b_i \rVert_2 > \epsilon.
\end{equation}
By \eqref{eq:example-uncovered-clique-l2-2}, we have that 
\begin{equation} \label{eq:example-uncovered-clique-l2-22}
    \sum_{i \in S} \lVert b_S - b_i \rVert_2^2 > m \epsilon^2.
\end{equation}
Now, from the assumption that the intersection is non-empty, take any  $\bar{x} \in \cap_{i \in S} B_2(b_i, \epsilon) \ne \emptyset$. Note that this implies that  
\begin{equation} \label{eq:example-uncovered-clique-l2-3}
    \forall i \in S, \quad \lVert \bar{x} - b_i \rVert_2^2\le  \epsilon^2.
\end{equation}
Recall that the average $b_S$ has the property of minimizing the sum of squared norms, \textit{i.e.}
\begin{equation} \label{eq:example-uncovered-clique-l2-centroid}
    b_s \in \argmin_{x \in \mathbb{R}^K} \sum_{i \in S} \lVert x - b_i \rVert_2^2.
\end{equation}
However, by \eqref{eq:example-uncovered-clique-l2-22} 
and \eqref{eq:example-uncovered-clique-l2-3} we get that  
\begin{equation*} \label{eq:example-uncovered-clique-l2-4}
    \sum_{i \in S} \lVert \bar{x} - b_i \rVert_2^2 \le m \epsilon^2 < \sum_{i \in S} \lVert b_S - b_i \rVert_2^2.
\end{equation*}
This contradicts \eqref{eq:example-uncovered-clique-l2-centroid}.
\end{example}

\begin{example}[Odd anti-holes with the $\ell_{\infty}$ norm] \label{example:anti-holes-existence}

The anti-hole of size 5 is isomorphic to the hole of size 5, which is always possible to build. Let us see that for sizes greater than 5, it remains possible to build odd anti-holes.

Fix $d = 7$, and consider the following points in $\mathbb{R}^7$:

\begin{equation*}
\begin{array}{lllllllll}
     x_1 = & (0   & ,0.2 & ,0.3 & ,0.4 & ,0.5 & ,0.6 & ,1   & ) \\
     x_2 = & (1   & ,0   & ,0.3 & ,0.4 & ,0.5 & ,0.6 & ,0.7 & ) \\
     x_3 = & (0.1 & ,1   & ,0   & ,0.4 & ,0.5 & ,0.6 & ,0.7 & ) \\
     x_4 = & (0.1 & ,0.2 & ,1   & ,0   & ,0.5 & ,0.6 & ,0.7 & ) \\
     x_5 = & (0.1 & ,0.2 & ,0.3 & ,1   & ,0   & ,0.6 & ,0.7 & ) \\
     x_6 = & (0.1 & ,0.2 & ,0.3 & ,0.4 & ,1   & ,0   & ,0.7 & ) \\
     x_7 = & (0.1 & ,0.2 & ,0.3 & ,0.4 & ,0.5 & ,1   & ,0   & ) \\
\end{array}
\end{equation*}

Let $\epsilon = 0.5 - 0.01$. Then it can be seen that the following properties hold:

\begin{itemize}
    \item $\forall i \in [7], B_\infty(x_i, \epsilon) \cap B_\infty(x_{(i-1)[\mathrm{mod} \, 7]}, \epsilon) = B_p(x_i, \epsilon) \cap B_p(x_{(i+1)[\mathrm{mod} \, 7]}, \epsilon) = \emptyset$
    \item $\forall i, j \in [7]  \text{ such that } i < j \text{ and } j - i \neq 1 [\mathrm{mod} \, 7],  \quad  B_p(x_i, \epsilon) \cap B_p(x_{j}, \epsilon) \ne \emptyset$
\end{itemize}

In other words, these points form an anti-hole of size 7. By an analogous reasoning, one can conclude that odd anti-holes of any size exist for a sufficiently large dimension $d$.
    
\end{example}


\section{SUPPLEMENTARY MATERIAL FOR SECTION~\ref{sec:rg is big}: THE RANDOMIZATION GAP CAN BE ARBITRARILY CLOSE TO \texorpdfstring{$\nicefrac{1}{2}$}{1/2}} \label{app: section 4 randomization gap arbitrarily large}

The proof of Theorem~\ref{thm:fibration-arbitrary-rg} relies on two key elements: (i) the construction of a non-perfect graph for which the value of the set packing problem and its fractional counterpart differ significantly, and (ii) the representation of this difference as the randomization gap of a distribution. In Section~\ref{subsec:every-graph-is-intersection}, we demonstrate through Lemma~\ref{lemma: build data set from graph G} that any loopless graph is isomorphic to the conflict graph of a distribution. Furthermore, if the graph is triangle-free, the representation in (ii) is amenable. In Section~\ref{app:section-5-iterative-construction}, we establish Corollary~\ref{cor:fibration-graph-arbitrary-rg}, which provides a triangle-free construction for (i), based on the iterative procedure presented in~\cite{chung1993note}.  Combining both results, we prove Theorem~\ref{thm:fibration-arbitrary-rg} in Section~\ref{app:section-5-proof-thm}.




\subsection{Every graph is the conflict graph of a distribution}
\label{subsec:every-graph-is-intersection}

In Lemmas~\ref{lemma: cubicity} and~\ref{lemma: sphericity general} we prove that for the $\ell_\infty$ and $\ell_p$ norms (with $p \in (1, \infty)$) respectively, any graph $G$ can be constructed by considering the overlap of some $\epsilon$-balls. This corresponds to the second condition in the definition of the conflict graph (Definition~\ref{def: conflict hypergraph}). Building on this, we utilize the chromatic number (Definition~\ref{def: chromatic number}) to restrict the overlap of $\epsilon$-balls to points from different classes (satisfying the first condition in Definition~\ref{def: conflict hypergraph}). Consequently, we prove Lemma~\ref{lemma: build data set from graph G}, which states that $G$ is isomorphic to the conflict graph $\confgraph$ of a distribution $\mu$. Furthermore, if $G$ is triangle-free, then it is isomorphic to the loopless version of the conflict hypergraph $\confhyp$. This implies that the values of their respective (fractional) set packing problems coincide.


Lemma~\ref{lemma: restatment intersecting neighborhoods} establishes a straightforward equivalence that will be used in the proofs throughout this subsection.

\begin{lemma}[Restatement of intersecting neighborhoods] \label{lemma: restatment intersecting neighborhoods}
Let us consider an $\ell_{p}$ norm with $p \in (1, \infty]$ and $\epsilon>0$. The following assertion holds true:

\begin{equation*}
   \forall x, x' \in \mathcal{X}, \quad  B_p(x, \epsilon) \cap B_{p}(x', \epsilon) \ne \emptyset \iff \lVert x - x' \rVert_{p} \le 2\epsilon
\end{equation*}
\end{lemma}
\begin{proof}
    First, suppose that $B_p(x, \epsilon) \cap B_p(x', \epsilon) \ne \emptyset$ and take any $z \in B_p(x, \epsilon) \cap B_p(x', \epsilon)$. Then, by the triangle inequality
    \begin{align*}
        \lVert x - x' \rVert_{p} & = \lVert x - z + z - x' \rVert_{p} \le \lVert x - z \rVert_{p} + \lVert z - x' \rVert_{p} \le \epsilon + \epsilon  = 2 \epsilon\\
    \end{align*}

    For the other direction, suppose that $\lVert x - x' \rVert_{p} \le 2 \epsilon$ and consider the point $z = \frac{x + x'}{2}$. Then, 
    \begin{align*}
        \lVert x - z \rVert_{p} & = \lVert x - \frac{x + x'}{2}\rVert_{p}  = \lVert \frac{x - x'}{2} \rVert_{p} = \frac{1}{2} \lVert x - x' \rVert_{p} \le \frac{1}{2} 2 \epsilon = \epsilon
    \end{align*}

    Thus, $z \in B_{p}(x, \epsilon)$. An analogous argument exchanging $x$ by $x'$ yields that $z \in B_{p}(x', \epsilon)$.
    As $z \in B_p(x, \epsilon) \cap B_p(x', \epsilon)$, we conclude that $B_p(x, \epsilon) \cap B_p(x', \epsilon) \ne \emptyset$. 
    
\end{proof}

Now we restate and prove lemmas related to the cubicity and sphericty of graphs. Basically, these results show for any $p$-norm, $\epsilon > 0$ and graph $G$, we can think of $G$ as the intersection graph of some set of $|G|$ points. This will be the first step in the construction of discrete distributions, as it provides the support of the distribution in such a way that the conflicts between points are exactly those represented in $G$.
\begin{lemma}[Cubicity, restated from~\cite{roberts1969boxicity}] \label{lemma: cubicity}
    Let $G=(V, E)$ be any graph with $n \in \mathbb{N}^*$ vertices. Let us consider the  $\ell_{\infty}$ norm and $\epsilon>0$. There exist $d \in \mathbb{N}^*$ and a set of points $ x_1, \dots, x_n$ from $\mathbb{R}^d$ such that  the following holds:
    \begin{equation*} \label{eq: adjacency cubicity equation}
    \forall i, j \in V, \quad \{i, j \} \in E \iff B_\infty( x_i, \epsilon) \cap B_\infty(x_j, \epsilon) \neq \emptyset
    \end{equation*}

\end{lemma}

\begin{proof}

Following the remark given in~\citep[Section 2]{roberts1969boxicity}, we build coordinate functions to embed the \( n \) vertices of \( G \) into \( \mathbb{R}^n \). For any \( i \in V \), consider the following vector \( x_i = (x_i^{(1)}, x_i^{(2)}, \dots, x_i^{(n)}) \in \mathbb{R}^n \), where:

\begin{equation*} \label{eq: coordinate functions for cubicity proof}
        \forall j \in V, ~ x_i^{(j)} = \begin{cases}
			0, & \text{if $j = i$}\\
                0.9 \epsilon, & \text{if $i \neq j$ and $\{i, j \} \in E$} \\
                1.1 \epsilon, & \text{if $i \neq j$ and $\{i, j \} \notin E$} \\
		 \end{cases}
    \end{equation*}

Now, we distinguish between two key cases.

\textbf{Case 1: If \( \{i, j \} \in E \), the distance \( \lVert x_i - x_j \rVert_{\infty} \) is strictly less than \( \epsilon \).} We need to verify that for all \( k \in [n] \), \( |x_i^{(k)} - x_j^{(k)}| < \epsilon \). Let us consider each coordinate \( k \):

- If \( k = i \), then:
  \[
  |x_i^{(k)} - x_j^{(k)}| = |0 - x_j^{(i)}| = x_j^{(i)} = 0.9 \epsilon,
  \]

- Similarly, if \( k = j \), then:
  \[
  |x_i^{(k)} - x_j^{(k)}| = |0 - x_i^{(j)}| = x_i^{(j)} = 0.9 \epsilon,
  \]

- If \( k \neq i \) and \( k \neq j \), then both \( x_i^{(k)} \) and \( x_j^{(k)} \) are either \( 0.9\epsilon \) or \( 1.1\epsilon \), depending on whether \( \{i,k\} \in E \) and \( \{j,k\} \in E \). In any case, we have:
  \[
  |x_i^{(k)} - x_j^{(k)}| \leq |1.1\epsilon - 0.9\epsilon| = 0.2\epsilon < \epsilon.
  \]

Thus, for all coordinates \( k \in [n] \), we have \( |x_i^{(k)} - x_j^{(k)}| < \epsilon \). Therefore, \( \lVert x_i - x_j \rVert_{\infty} < \epsilon \), as desired.

\textbf{Case 2: If \( \{i, j \} \notin E \), the distance \( \lVert x_i - x_j \rVert_{\infty} \) is strictly greater than \( \epsilon \).}

- If \( k = i \) or \( k = j \), then:
  \[
  |x_i^{(k)} - x_j^{(k)}| = 1.1 \epsilon > \epsilon
  \]

Thus,  if \( \{i, j \} \notin E \) then \( \lVert x_i - x_j \rVert_{\infty} > \epsilon \).

\textbf{Conclusion.} We have shown that \( \lVert x_i - x_j \rVert_{\infty} \leq \epsilon \) if and only if \( \{i, j \} \in E \).

By replacing \( \epsilon \) by \( 2 \epsilon \) and using Lemma~\ref{lemma: restatment intersecting neighborhoods}, we obtain that:

\begin{equation*} \label{eq: final proof cubicity}
     \{i, j \} \in E  \iff B_\infty(x_i, \epsilon) \cap B_\infty(x_j, \epsilon) \neq \emptyset. 
\end{equation*}

This concludes the proof.

\end{proof}

We now prove a result related to the concept of 
sphericity \cite{fishburn1983sphericity} of a graph, but generalized to any $p$-norm with $p \in (1, \infty)$.

\addtocounter{lemma}{1}
\begin{lemma}[Sphericity for any $p$-norm] \label{lemma: sphericity general}
     Let $G=(V, E)$ be any graph with $n \in \mathbb{N}^*$ vertices. Let us consider an $\ell_{p}$ norm with $p \in (1, \infty)$ and $\epsilon>0$. There exist $d \in \mathbb{N}^*$ and a set of points $x_1, \dots, x_n$ from $\mathbb{R}^d$ such that for any $i,j \in V$ such that $i \neq j$, the following holds:
    \begin{equation} \label{eq: adjacency sphericity general equation}
    \{i, j \} \in E \iff B_p(x_i, \epsilon) \cap B_p(x_j, \epsilon) \neq \emptyset.
    \end{equation}

\end{lemma}

\begin{proof} Let $G=(V, E)$ be any graph with $n \in \mathbb{N}^*$ vertices and $m \in \left[\frac{ (n-1)n}{2} \right] $ edges. For any $i \in V$, we denote by $\mathrm{deg}_{i}$ the degree of $i$, i.e., the number of neighbors of $i$ in $G$. Set $d = n+m$, and define $x_1, \dots x_n \in \mathbb{R}^{n+m}$ as follows. 

\textbf{Definition of the vectors.} For any $i \in [n]$ and $k \in [d]$, define $x_i^{(k)}$ as 
\begin{equation*} 
        x_i^{(k)} = \begin{cases}
			1, & \text{if $k \in [m]$ and $i \in e_k$}\\
            \left(n-\mathrm{deg}_{i}\right)^{\frac{1}{p}}, & \text{if $k =m+i$} \\
            0, & \text{otherwise}.   \\
		 \end{cases}
    \end{equation*}
Let us now consider $i,j \in V$ such that $i \neq j$, we can distinguish two cases.

\textbf{Case 1.} If $\{ i,j \} \notin E$ then $\nexists k \in [m]$ such that $x_i^{(k)}= x_j^{(k)}=1$. Furthermore, as $i \neq j$, we have $m+i \neq m +j$. Hence, we have $\left\Vert x_{i}-x_{j}\right\Vert _{p}^ {}=\left( \mathrm{deg}_{i}+\mathrm{deg}_{j}+\left(n-\mathrm{deg}_{i}\right)+\left(n-\mathrm{deg}_{j}\right)\right) ^{\frac{1}{p}}=\left( 2n\right) ^{\frac{1}{p}}$. \\

\textbf{Case 2.} If $\{i,j\}\in E$, then $\exists! k \in [m]$ such that $x_i^{(k)}= x_j^{(k)}=1$. Furthermore, $m+i \neq m +j$. Hence, we have $\left\Vert x_{i}-x_{j}\right\Vert _{p}^ {}=\left( \mathrm{deg}_{i}+\mathrm{deg}_{j}-1+\left(n-\mathrm{deg}_{i}\right)+\left(n-\mathrm{deg}_{j}\right)\right)^{\frac{1}{p}}=\left( 2n-1\right) ^{\frac{1}{p}}.$

Finally, multiplying each $x_{i}$ by a constant $2\epsilon\cdot\left( 2n-1\right) ^{-\frac{1}{p}}$
we obtain that
\begin{equation*} 
        \left\Vert x_{i}-x_{j}\right\Vert _{p} = \begin{cases}
			2\epsilon \cdot \left( \frac{2n}{2n - 1} \right)^{\frac{1}{p}}, & \text{if $\{i, j\} \notin E$}\\
            2\epsilon 
            , & \text{if $\{i, j\} \in E$}
		 \end{cases}
\end{equation*}
As $\left( \frac{2n}{2n - 1} \right)^{\frac{1}{p}} \geq 1$ for any $n \in \mathbb{N}^*$ and $p \in (1, \infty)$, we finally get that 
\begin{equation*}
    \left\Vert x_{i}-x_{j}\right\Vert _{p} \leq 2\epsilon \iff \{i, j\} \in E.
\end{equation*}
By Lemma \ref{lemma: restatment intersecting neighborhoods}, we conclude that \eqref{eq: adjacency sphericity general equation} holds.
\end{proof}

Now that we know that for any graph $G$ we can find a set of points $S$ such that $G \simeq \confgraph[S]$, we only need to include the labels of the points to be able to create a discrete distribution for a classification task. Interestingly, the condition that all points in conflict must be from different classes has a direct counterpart in graph theory: \emph{vertex colorings}. We define below the chromatic number before proving Lemma~\ref{lemma: build data set from graph G}.

\begin{definition}[Chromatic number $\chi(G)$] \label{def: chromatic number}
    Given a graph $G = (V, E)$, a \emph{coloring} of the vertices of $G$ is a function $c: V \to \mathbb{N}$ such that 
    \begin{equation*}
        \forall~ \{i, j \} \in E, \quad c(i) \neq c(j). 
    \end{equation*} 
    The \emph{chromatic number} of $G$, denoted $\chi(G)$, is the smallest number of colors needed to color the vertices of $G$. That is, $\chi(G)$ is the smallest $M$ such that there exists a coloring $c: V \to [M]$.
\end{definition}


\begin{restatable}{lemma}{lemmaBuildDatasetFromGraph} \label{lemma: build data set from graph G}
    Let $G=(V, E)$ be any loopless graph with $n \in \mathbb{N}^*$ vertices. Let us consider an $\ell_{p}$ norm with $p \in (1, \infty]$ and $\epsilon>0$. There exist $d, K \in \mathbb{N}$ and $S = \{(x_i, y_i)\}_{i \in [n]} \subset \mathbb{R}^d \times [K]$ , such that $\confgraph[S] \simeq G$,
    where the relation $\simeq$ denotes graph isomorphism. Moreover, if $G$ is triangle-free, then $G$ is isomorphic to the loopless version of $\confhyp[S]$ and the number of classes satisfies $K \in \cal O\left(\sqrt{\nicefrac{n}{\log n}}\right)$, where $n = |S|$.
\end{restatable}
\begin{proof}
    Using either Lemma~\ref{lemma: cubicity} or Lemma~\ref{lemma: sphericity general} for $p = \infty$ or $p\in (1, \infty)$ respectively, there exist $d \in \mathbb{N}^*$ and a set of points $\{x_i\}_{i \in [n]}$ in $\mathbb{R}^d$ such that for any $i, j \in V$ with $ i\neq j$, the following holds: 

    \begin{equation} \label{eq: proof lemma graph embedding edge condition}
        \{i, j \} \in E \iff B_p(x_i, \epsilon) \cap B_p(x_j, \epsilon) \ne \emptyset.
    \end{equation}

    For the number of classes, take $K = \chi(G)$, the chromatic number of $G$. By definition of the chromatic number, there exists a $K$-coloring of $G$, \textit{i.e} a function $c: V \mapsto [K]$, such that for any $i,j \in V$
    \begin{equation} \label{eq: coloring of G}
        \{i, j\} \in E \implies c(i) \ne c(j).
    \end{equation}
    We define $S = \{(x_i, c(i))\}_{i \in [n]}$ and show that $\confgraph[S]$ is isomorphic to $G$. 

    {\bf Proving isomorphism.}
    As $G$ and $\confgraph[S]$ share the same set of vertices, we have to show that $$\{i, j \} \in E \iff \{i, j \} \in E(\confgraph[S]).$$
    If $\{i, j \} \in E$, as $G$ is loopless we have $i \neq j$, and therefore by \eqref{eq: proof lemma graph embedding edge condition} and \eqref{eq: coloring of G} we know that $ B_p(x_i, \epsilon) \cap B_p(x_j, \epsilon) \ne \emptyset $ and  $ c(i) \ne c(j)$. These two conditions imply that $\{i, j\} \in E(\confgraph[S])$ by definition of the conflict graph.

    On the other hand, if $ \{i,j\} \in E(\confgraph[S])$, we have that in particular, the $\epsilon$-balls intersect, \textit{i.e.} $B_p(x_i, \epsilon) \cap B_p(x_j, \epsilon) \ne \emptyset $. As $\confgraph[S]$ is loopless, we have $i \neq j$, and this implies that $\{i, j \} \in E $ by \eqref{eq: proof lemma graph embedding edge condition}.

    {\bf Adding the triangle-free assumption.}
    Note that we do not know the full structure of the conflict hypergraph $\confhyp[S]$, apart from the fact that the corresponding conflict graph $\confgraph[S]$ is isomorphic to the graph $G$. However, if we add the assumption that $G$ is triangle-free, we can show that $\confhyp[S]$ is the loopless version of $G$. Suppose by contradiction that $\confhyp[S]$ has a hyperedge $e \in \mathcal{E}(\confhyp[S])$ of size greater than 2. Then there would be a triplet of points $x_i, x_j, x_k$ such that 
    $$B_p(x_i, \epsilon) \cap B_p(x_j, \epsilon) \cap B_p(x_k, \epsilon) \ne \emptyset.$$
    This would imply the existence of the three edges $\{i, j \}, \{i,k\}$ and $\{j,k\}$ in  $\confgraph[S]$, and therefore in $G$, which form a triangle. As $G$ is triangle-free, we can conclude that $\confhyp[S]$ cannot contain any hyperedge of size greater than 2. Thus, by definition of conflict graph $\confgraph[S]$, we conclude that $G$ is isomorphic to the loopless version of $\confhyp[S]$.

    Regarding the number of classes when $G$ is triangle-free, we know that $K = \chi(G) \in \cal O \left( \sqrt{\frac{n}{\log n}} \right)$ (See \cite{davies2022chi,erdos1985chromatic,jensen2011graph}).
    
\end{proof}



\subsection{Constructing non-perfect graphs with large randomization gap}
\label{app:section-5-iterative-construction}

Lemma~\ref{lemma: build data set from graph G} ensures that any graph $G$ can be materialized as the conflict graph of some discrete distribution $\mu$. Moreover, if the $G$ is triangle-free, then $G$ is essentially the conflict hypergraph, which implies that we can compute the randomization gap of $\mu$ using only the (fractional) set packing problem over $G$. To prove Theorem~\ref{thm:fibration-arbitrary-rg} we thus only need to build a graph $G$ for which we can control the gap between the value of the fractional set packing problem and the set packing problem. With this objective in mind, we go over the construction proposed in \cite{chung1993note} of a graph that is triangle-free and without large independent sets. We restate and adapt their main results to our notation for completeness.
The construction in \cite{chung1993note} is iterative. Starting from a graph $G$, the authors propose to build a new, larger graph $H$ called its \emph{fibration}, using 6 copies of $G$ and connecting vertices between copies in a particular manner. This construction will preserve the property of being triangle-free, will increase the number of nodes by a factor of 6, but more importantly, it will at most increase the size of the largest independent set by a factor of 4. Let us formalize this.

\begin{definition}[Independence number $\alpha(G)$] \label{def: independence number}
    Given a graph $G$, the \emph{independence number} of $G$, denoted $\alpha(G)$, is the size of a maximum independent set of $G$.
\end{definition}

\begin{definition}[Fibration \cite{chung1993note}]
    For any graph $G = (V(G), E(G))$, the \emph{fibration} of $G$ is the graph $H = (V(H), E(H))$ defined as follows:
    \begin{enumerate}
        \item $V(H) = V(G) \times [6].$
        \item $\forall i \in [6], \quad \{u, v\} \in E(G) \implies \{(u, i), (v, i)\} \in E(H).$
        \item $\forall i, j \in [6] \text{ with } j \equiv (i+1)[\mathrm{mod}~6], \quad \{u, v\} \in E(G) \implies \{(u, i), (v, j)\} \in E(H) \text{ and } \{(u, j), (v, i)\} \in E(H).$
        \item $\forall i, j \in [6] \text{ with } j \equiv (i+3)[\mathrm{mod}~6], \quad u \in V(G) \implies \{(u, i), (u, j)\} \in E(H).$
    \end{enumerate}

\end{definition}

In Figure \ref{fig:fibration-graph-appendix} we show an example of the fibration of the cycle with three nodes $C_3$.
There are two properties of the fibration that are important:

\begin{lemma}[Triangle-free property preservation, Lemma 1 in \cite{chung1993note}]\label{lemma:chung1}
    If $G$ is triangle-free and $H$ is the fibration of $G$, then $H$ is triangle-free.
\end{lemma}

\begin{lemma}[Bound on the maximum independent set, Lemma 2 in  \cite{chung1993note}] \label{lemma:chung2}
    If $H$ is the fibration of $G$, then $\alpha(H) \le 4 \cdot \alpha(G)$.
\end{lemma}

\begin{figure}
    \centering
    \begin{tikzpicture}[scale = 2]
        \def\radius{4}

        \begin{scope}[scale=2, every node/.style={fill=white, inner sep=0pt}]
        \foreach \i in {1,2,...,6} {
            \node (Z\i) at ({60*(\i-1)}:1) {};
        }

        \end{scope}

        \begin{scope}[scale=2, every node/.style={circle, draw, fill=white, inner sep=5pt}]
        
        \foreach \i/\j in {4/5, 5/6, 6/1} {
            \node (A\i) at ($(Z\i)!0.20!(Z\j)$) {\scriptsize $a$};
            \node (B\i) at ($(Z\i)!0.50!(Z\j)$) {\scriptsize $b$};
            \node (C\i) at ($(Z\i)!0.80!(Z\j)$) {\scriptsize $c$};
        }
        
        \foreach \i/\j in {1/2, 2/3, 3/4} {
            \node (A\i) at ($(Z\i)!0.20!(Z\j)$) {\scriptsize $a$};
            \node (B\i) at ($(Z\i)!0.50!(Z\j)$) {\scriptsize $b$};
            \node (C\i) at ($(Z\i)!0.80!(Z\j)$) {\scriptsize $c$};
        }

        \end{scope}

        \begin{scope}[color=black]
        \foreach \i/\j in {1/2, 2/3, 3/4} {
            \draw (A\i) -- (B\i) -- (C\i);
            \draw (A\i) to[bend right=60] (C\i);
        }
        \end{scope}

        \begin{scope}[color=black]
        \foreach \i/\j in {4/5, 5/6, 6/1} {
            \draw (A\i) -- (B\i) -- (C\i);
            \draw (A\i) to[bend right=60] (C\i);
        }
        \end{scope}

        \begin{scope}[color=red, dashed]
        \foreach \i/\j in {1/2, 2/3, 3/4, 4/5, 5/6, 6/1} {
            \draw (A\i) to[bend left=30] (B\j);
            \draw (B\i) to[bend left=50] (A\j);
            \draw (B\i) to[bend left=30] (C\j);
            \draw (C\i) to[bend left=50] (B\j);
            \draw (A\i) to[bend left=40] (C\j);
            \draw (C\i) to[bend left=0] (A\j);
        }
        \end{scope}

        \begin{scope}[color=blue]
        \foreach \i/\j in {1/4, 2/5, 3/6} {
            \draw (A\i) to[bend left=0] (A\j);
            \draw (B\i) to[bend left=0] (B\j);
            \draw (C\i) to[bend left=0] (C\j);
        }
        \end{scope}

    \end{tikzpicture}
    \caption{Example of the graph $G_1$ built in \cite{chung1993note} when the initial $G_0$ is the 3-cycle $C_3$. There are 6 copies of $C_3$ with nodes labeled $a,b$ and $c$. Black edges are those within each copy, while \textcolor{red}{red} and \textcolor{blue}{blue} edges are the ones added by the construction in \cite{chung1993note} between different copies of the initial graph.}
    \label{fig:fibration-graph-appendix}
\end{figure}

To construct our graph of interest, we will iteratively apply the fibration operation, starting from an initial graph. Let $G_0$ be a fixed triangle-free graph, and let $G_t$ denote the graph obtained after applying the fibration operation $t$ times to $G_0$. Then, by Lemmas \ref{lemma:chung1} and \ref{lemma:chung2}, we have the following properties:
\begin{align}
    & G_t \text{ is triangle-free}  \nonumber \\
    & |G_t| = 6^t \cdot |G_0| \label{eq:nodes-of-fibration}\\
    & \alpha(G_t) \le 4^t \cdot \alpha(G_0) \label{eq:alphagt}
\end{align}

For any graph $G$, the set packing problem with the weight vector $\vecones[|G|]$ has a natural interpretation: it is exactly the size of the maximum independent set. In other words, for any graph $G$, 
\begin{equation} \label{eq:ip-is-alpha}
    \mathrm{IP}(G, \vecones[|G|]) = \alpha(G).
\end{equation}

Coming back to the graph $G_t$, by  \eqref{eq:alphagt} and \eqref{eq:ip-is-alpha}, we can conclude that 
\begin{equation} \label{eq:chung:alpha}
    \mathrm{IP}(G_t, \vecones[|G_{t}|]) \le 4^t \alpha(G_0).
\end{equation}

On the other hand, for any graph $G$, the vector $\frac{1}{2}\vecones[|G|]$ is feasible for the fractional set packing problem because there are no hyperedges of size larger than 2, so any constraint in the definition of fractional packing involves at most 2 vertices. In other words, for any graph $G$,
\begin{equation} \label{eq:fp-half-isfeasible}
    \mathrm{FP}(G, \vecones[|G|]) \ge \frac{1}{2}\vecones[|G|]^T \vecones[|G|]  \ge \frac{|G|}{2}.
\end{equation}

Now let us consider the normalized vector $\unifvec[|G_t|]$. Clearly, we have that
\begin{align}
    & \mathrm{IP}(G_t, \unifvec[|G_t|]) = \frac{1}{|G_t|} \mathrm{IP}(G_t, \vecones[|G_t|]) \label{eq:ip-chung-section-1} \\
    & \mathrm{FP}(G_t, \unifvec[|G_t|]) = \frac{1}{|G_t|} \mathrm{FP}(G_t, \vecones[|G_t|]) \label{eq:fp-chung-section-1}
\end{align}

By \eqref{eq:fp-half-isfeasible} and \eqref{eq:fp-chung-section-1}, we can lower bound the value of the fractional set packing problem over $G_t$ as follows: 
\begin{equation} \label{eq:fp-chung-section-2}
    \mathrm{FP}(G_t, \unifvec[|G_t|]) \ge \frac{1}{2}.
\end{equation}

Now we upper bound the value of the set packing problem for the $t$-fibration graph:
\begin{align}
    \mathrm{IP}(G_t, \unifvec[|G_t|]) & = \frac{1}{|G_t|} \mathrm{IP}(G_t, \vecones[|G_t|]) & (\text{By \eqref{eq:ip-chung-section-1}})\\
    & \le  \frac{4^t}{|G_t|} \alpha(G_0) & (\text{By \eqref{eq:chung:alpha}}) \\
    & \le  \frac{4^t}{6^t} \alpha(G_0) & (\text{By \eqref{eq:nodes-of-fibration}}) \label{eq:chung:ip-upper-bound}
\end{align}

From \eqref{eq:fp-chung-section-2} and \eqref{eq:chung:ip-upper-bound} we can immediately deduce the following:

\begin{corollary}\label{cor:fibration-graph-arbitrary-rg}
For any graph $G_0$, let $G_t$ be the graph obtained by applying the fibration operation $t$ times, starting from $G_0$. Then, for any $\delta>0$, the following holds:
\begin{equation} \label{eq:cor:fibration-graph-arbitrary-rg}
    t > \displaystyle\frac{\log(\nicefrac{\delta}{\alpha(G_0)})}{\log(\nicefrac{2}{3})} \implies \mathrm{FP}(G_t, \unifvec[|G_t|]) - \mathrm{IP}(G_t, \unifvec[|G_t|]) \ge \frac{1}{2} - \delta
\end{equation}
\end{corollary}
\begin{proof}

    If $t>\frac{\log\left(\frac{\delta}{\alpha(G_{0})}\right)}{\log\left(\frac{2}{3}\right)}$
then we have
\begin{align*}
t\log\frac{2}{3} & \le\log\frac{\delta}{\alpha(G_{0})}\\
\left(\frac{2}{3}\right)^{t} & \le\frac{\delta}{\alpha(G_{0})}\\
\left(\frac{2}{3}\right)^{t}\alpha(G_{0}) & \le\delta
\end{align*}

Thus, with \eqref{eq:chung:ip-upper-bound}, $t>\frac{\log\left(\frac{\delta}{\alpha(G_{0})}\right)}{\log\left(\frac{2}{3}\right)}$
implies 
    $\mathrm{IP}(G_t, \unifvec[|G_t|]) \le \delta$. Then, injecting \eqref{eq:fp-chung-section-2}, we obtain the desired result.
\end{proof}

\subsection{Proof of Theorem~\ref{thm:fibration-arbitrary-rg}}
\label{app:section-5-proof-thm}

We are ready to state the main result of the section, that will use the fibration operation with Lemma \ref{lemma: build data set from graph G} to produce discrete distributions with randomization gap arbitrarily close to $\nicefrac{1}{2}$.

\thmChungDistribution*
\begin{proof}
    For the fixed $\epsilon$, $p$-norm and $\delta$, take any triangle-free loopless graph $G$ (for example, the 5-cycle $C_5$). Take any $t$ that satisfies the condition in \eqref{eq:cor:fibration-graph-arbitrary-rg} from Corollary \ref{cor:fibration-graph-arbitrary-rg} and consider $G_t$ the $t$-fibration of $G$.

    By Lemma \ref{lemma:chung1}, $G_t$ is triangle-free, and by Corollary \ref{cor:fibration-graph-arbitrary-rg}, we have that 
    \begin{equation} \label{eq:proof:main:fibration-for-dataset}
        \mathrm{FP}(G_t, \unifvec[|G_t|]) - \mathrm{IP}(G_t, \unifvec[|G_t|]) \ge \frac{1}{2} - \delta
    \end{equation}
   By Lemma \ref{lemma: build data set from graph G}, as $G_t$ is loopless, there exist $d, K \in \mathbb{N}$ and $S = \{(x_i, y_i)\}_{i \in [|G_t|]} \subset \mathbb{R}^d \times [K]$ such that $\confgraph[S] \simeq G_t$. Moreover, as $G_t$ is triangle-free, we have that $\confhyp[S]$ without considering loops is isomorphic to $G_t$. This implies\footnote{The constraints associated to loops are not considered in the definition of a fractional packing as $\mathcal{Q}^\mathrm{frac}(\confhyp[S]) \subseteq [0,1]^n$.} that 
    \begin{equation*}
        \forall \boldsymbol{\omega} \in \mathbb{R}_+^n, \quad \mathrm{FP}(\confhyp[S], \boldsymbol{\omega}) = \mathrm{FP}(\confgraph[S], \boldsymbol{\omega}). 
    \end{equation*}
    Define $\mu$ as the discrete distribution with support $\suppmu = S$ and uniform probability vector $ \omegamu = \unifvec[|G_t|]$.  
    Given that $\mathrm{IP}(\confhyp, \omegamu) = \mathrm{IP}(\confgraph, \omegamu)$ by Corollary~\ref{lemma: reformulations of MIS}, we can conclude thanks to \eqref{eq:proof:main:fibration-for-dataset} that 
    \begin{equation*}
        \mathrm{rg}(\mu, \epsilon) = \mathrm{FP}(\confhyp, \omegamu) - \mathrm{IP}(\confhyp, \omegamu) = \mathrm{FP}(\confgraph, \omegamu) - \mathrm{IP}(\confgraph, \omegamu) \ge \frac{1}{2} - \delta.
    \end{equation*}

    To check that $\confhyp$ is conformal, notice that as $\confgraph$ is triangle-free, the largest cliques of $\confgraph$ are the 2-edges, which are also hyperedges in $\confhyp$. Thus, all cliques from $\confgraph$, including the maximal ones, are hyperedges of  $\confhyp$. Lastly, by Theorem \ref{thm: optimal classifier is deterministic}, that $\confgraph$ is not perfect. 

    The fact that $K \in \cal O(\sqrt{\frac{n}{\log n}})$ where $n = |G_t|$ is also given by Lemma \ref{lemma: build data set from graph G} because $G_t$ is triangle-free.
    
\end{proof}

\setcounter{section}{4} 
\section{SUPPLEMENTARY MATERIAL: ON THE HARDNESS OF COMPUTING THE RANDOMIZATION GAP.}\label{app:hardness-section}

Here, we formally redefine the optimization and decision problems discussed in the paper. Our focus is to express these problems in the context of hypergraph-based set packing, providing clear formulations for both their optimization and decision versions. We then prove the co-NP-completeness of the Randomization Gap decision problem.

\subsection{Definition of the Optimization Problems}

\subsubsection*{Optimization Problem: Weighted Set Packing $\mathrm{IP}\left(H,\boldsymbol{\omega}\right)$}

\hspace{24pt} \KwIn{A hypergraph \( H = (V, \mathcal{E}) \) and a weight vector $\boldsymbol{\omega} \in \mathbb{Q}_+^{|V|}$}
\hspace{24pt} \KwOut{ $\sup_{q \in \mathcal{Q}(H)} \boldsymbol{\omega}^T q$}

\vspace{1cm}

\subsubsection*{Optimization Problem: \emph{Fractional} Weighted Set Packing $\mathrm{FP}\left(H,\boldsymbol{\omega}\right)$}

\hspace{24pt} \KwIn{A hypergraph \( H = (V, \mathcal{E}) \) and a weight vector $\boldsymbol{\omega} \in \mathbb{Q}_+^{|V|}$}
\hspace{24pt} \KwOut{$\sup_{q \in ConvexHull\left(\mathcal{Q}(H)\right)} \boldsymbol{\omega}^T q$}
\vspace{1cm}

\subsection{Decision Problems Associated to these Optimization Problems}
\subsubsection*{Decision Problem: Weighted Set Packing $\mathrm{IP}\left(H,\boldsymbol{\omega},\alpha\right)$}

\hspace{24pt} \KwIn{A hypergraph \( H = (V, \mathcal{E}) \) and a weight vector \( \boldsymbol{\omega} \in \mathbb{Q}_+^{|V|} \), a rational number \( \alpha \in \mathbb{Q} \)}
\hspace{24pt} \KwOut{\texttt{YES} if there exists a packing \( q \in \mathcal{Q}(H) \) such that \( \boldsymbol{\omega}^T q \geq \alpha \), otherwise \texttt{NO}}
\vspace{1cm}

\subsubsection*{Decision Problem: Randomization Gap Problem $\mathrm{RG}\left(H,\boldsymbol{\omega},\alpha\right)$}

\hspace{24pt} \KwIn{A hypergraph \( H = (V, \mathcal{E}) \) and a weight vector \( \boldsymbol{\omega} \in \mathbb{Q}_+^{|V|} \), a rational number \( \alpha \in \mathbb{Q} \)}
\hspace{24pt} \KwOut{\texttt{YES} if and only if \( \mathrm{FP}(H, \boldsymbol{\omega}) - \mathrm{IP}(H, \boldsymbol{\omega}) \geq \alpha \), otherwise \texttt{NO}}
\vspace{1cm}

\subsection{Reminder on NP-hardness and co-NP-hardness}

Only \emph{decision} problems are in \textbf{NP} or in \textbf{co-NP}.

\begin{itemize}
    \item A decision problem is in \textbf{NP} if, given a "yes" instance\footnote{a YES instance is a input on which a correct algorithm should output YES} \( x \), there exists a certificate (or witness) \( c \) that can be verified in polynomial time. 
    Formally, the problem is in \textbf{NP} is there exists a polynomial-time verification algorithm \( A(x, c) \) such that:
    \begin{itemize}
        \item If \( x \) is a "yes" instance, then there exists a certificate \( c \) such that \( A(x, c) \) outputs \texttt{yes}.
        \item Reciprocally, if \( x \) is a "no" instance, then for any certificate \( c \), \( A(x, c) \) will output \texttt{no}.
    \end{itemize}

    \item A decision problem is in \textbf{co-NP} if, given a "no" instance \( x \), there exists a certificate \( c \) that can be verified in polynomial time. 
    Formally, the problem is in \textbf{co-NP}  iff there exists a polynomial-time verification algorithm \( A(x, c) \) such that:
    \begin{itemize}
        \item If \( x \) is a "no" instance, then there exists a certificate \( c \) such that \( A(x, c) \) outputs \texttt{yes}.
        \item Reciprocally, if \( x \) is a "yes" instance, then for any certificate \( c \), \( A(x, c) \) will output \texttt{no}.
    \end{itemize}
\end{itemize}

A decision problem is NP-complete if it belongs to the class NP \emph{and} it is NP-hard. Same for co-NP.

\subsection{Hardness of the RG decision problem}

\begin{theorem}
The $\mathrm{RG}\left(H,\boldsymbol{\omega},\alpha\right)$ problem is co-NP complete
\end{theorem}
\begin{proof}
To show a problem is co-NP complete, we must first show it belongs to co-NP, and then that it is co-NP hard.

\textbf{The problem belongs to co-NP.}
To show it belongs to co-NP, we must show that there exist certificates for all NO instances.
For all NO instances $(H,\boldsymbol{\omega},\alpha)$  we have \( \mathrm{FP}(H, \boldsymbol{\omega}) -\alpha <  \mathrm{IP}(H, \boldsymbol{\omega})\). Here, there must exist a packing $q$  (which will be our certificated) such that $\mathrm{FP}\left(H,\boldsymbol{\omega}\right) - \alpha < \boldsymbol{\omega}^T q$. So for any NO instance, it suffices to pick any packing $q$ satisfying this inequality to convince a verification algorithm that our instance is in fact a NO instance. Thus, the problem belongs to co-NP.

\textbf{The problem is co-NP hard.}
To show it is co-NP-hard, we build a trivial reduction from the $\mathrm{IP}(H, \boldsymbol{\omega},\alpha)$  problem.
Assume by contradiction that there exists a polynomial time algorithm $Alg\left(H,\boldsymbol{\omega},\alpha\right)$ able to solve the $\text{RG}$ problem. Let us show that this would immediately imply the existence of a polynomial time algorithm to solve the $\mathrm{IP}$ problem, which  contradicts the well known NP-hardness of $\mathrm{IP}$. More precisely,  
Let $D$ be the least common denominator of $\boldsymbol{\omega}_1 \ldots \boldsymbol{\omega}_{\left| V \right|}$ and define the function $floor(z)=\frac{1}{D}\left\lfloor z\times D\right\rfloor$. 
Let $\left(H, \boldsymbol{\omega},\alpha'\right)$ be an arbitrary instance of the  $\mathrm{IP}$ problem. Define $\alpha=\mathrm{FP}\left(H,\boldsymbol{\omega}\right)-floor(\alpha')$. Observe that these conditions are equivalent:
\[
\begin{array}{c}
\mathrm{IP}(H, \boldsymbol{\omega}, \alpha') = \text{No} \\
\mathrm{IP}(H, \boldsymbol{\omega}) < \alpha' \\
\mathrm{IP}(H, \boldsymbol{\omega}) \le floor(\alpha') \\
\mathrm{FP}(H, \boldsymbol{\omega}) - \mathrm{IP}(H, \boldsymbol{\omega}) \ge \mathrm{FP}(H, \boldsymbol{\omega}) - floor(\alpha') \\
\mathrm{RG}\left(H,\boldsymbol{\omega},\alpha\right)=Yes
\end{array}
\]
Thus, we can run $Alg\left(H,\boldsymbol{\omega},\alpha\right)$ 
in polynomial time and return the opposite of the boolean answer to solve $\mathrm{IP}(H, \boldsymbol{\omega}, \alpha')$ also in polynomial time. This contradicts NP-hardness of $\mathrm{IP}$. Thus, our problem is co-NP-Hard, thus co-NP-complete.


\end{proof}

\end{document}

%% file: checklist.tex
\section*{Checklist}



 \begin{enumerate}

 \item For all models and algorithms presented, check if you include:
 \begin{enumerate}
   \item A clear description of the mathematical setting, assumptions, algorithm, and/or model. \textbf{[Yes]} We do not include algorithms. Our mathematical framework is clearly explained in the Preliminaries section, and all our Theorems and Lemmas also state all the assumptions.
   \item An analysis of the properties and complexity (time, space, sample size) of any algorithm. \textbf{[Yes]} Even if we do not have algorithms, we discuss the complexity of our problem and discuss why it is hard to find sufficient conditions in Section \ref{sec: sufficient conditions}. 
   \item (Optional) Anonymized source code, with specification of all dependencies, including external libraries. \textbf{[Not Applicable]} We do not have experiments nor source code.
 \end{enumerate}

 \item For any theoretical claim, check if you include:
 \begin{enumerate}
   \item Statements of the full set of assumptions of all theoretical results. \textbf{[Yes]}
   \item Complete proofs of all theoretical results. \textbf{[Yes]} All of them are in the Appendix to respect the space constraints. Most results used from previous work are restated, and always cited.
   \item Clear explanations of any assumptions. \textbf{[Yes]} We do our best to explain both intuitively and rigorously all the assumptions and implications of all our results.     
 \end{enumerate}

 \item For all figures and tables that present empirical results, check if you include:
 \begin{enumerate}
   \item The code, data, and instructions needed to reproduce the main experimental results (either in the supplemental material or as a URL). \textbf{[Not Applicable]}
   \item All the training details (e.g., data splits, hyperparameters, how they were chosen). \textbf{[Not Applicable]}
         \item A clear definition of the specific measure or statistics and error bars (e.g., with respect to the random seed after running experiments multiple times). \textbf{[Not Applicable]}
         \item A description of the computing infrastructure used. (e.g., type of GPUs, internal cluster, or cloud provider). \textbf{[Not Applicable]}
 \end{enumerate}

 \item If you are using existing assets (e.g., code, data, models) or curating/releasing new assets, check if you include:
 \begin{enumerate}
   \item Citations of the creator If your work uses existing assets. \textbf{[Not Applicable]}
   \item The license information of the assets, if applicable. \textbf{[Not Applicable]}
   \item New assets either in the supplemental material or as a URL, if applicable. \textbf{[Not Applicable]}
   \item Information about consent from data providers/curators. \textbf{[Not Applicable]}
   \item Discussion of sensible content if applicable, e.g., personally identifiable information or offensive content. \textbf{[Not Applicable]}
 \end{enumerate}

 \item If you used crowdsourcing or conducted research with human subjects, check if you include:
 \begin{enumerate}
   \item The full text of instructions given to participants and screenshots. \textbf{[Not Applicable]}
   \item Descriptions of potential participant risks, with links to Institutional Review Board (IRB) approvals if applicable. \textbf{[Not Applicable]}
   \item The estimated hourly wage paid to participants and the total amount spent on participant compensation. \textbf{[Not Applicable]}
 \end{enumerate}

 \end{enumerate}